\documentclass{jair}

\makeatletter
\renewcommand\footnotetextcopyrightpermission[1]{} 
\makeatother

\usepackage{graphicx}
\usepackage{caption}      
\usepackage{float}      
\usepackage[utf8]{inputenc}

\usepackage{amsmath,amssymb}
\usepackage{algorithm}
\usepackage{algpseudocode}
\usepackage{setspace}
\usepackage{wrapfig}
\usepackage{booktabs}   
\usepackage{siunitx}  

\usepackage{footmisc}     

\usepackage[normalem]{ulem}  
\newcommand{\largest}[1]{\dashuline{#1}}    
\newcommand{\smallest}[1]{\dotuline{#1}}   

\algrenewcommand{\algorithmicindent}{1.5em}        
\algrenewcommand{\algorithmiccomment}[1]{\hfill\(\#\) #1}

\usepackage{booktabs,siunitx,multirow,subcaption}
\renewcommand{\arraystretch}{0.9}
\begin{document}


\title{Explainability-Based Token Replacement on LLM-Generated Text}

\author{Hadi Mohammadi}
\authornote{Corresponding Author.}
\email{h.mohammadi@uu.nl}
\orcid{0000-0003-0860-9200}
\affiliation{%
  \institution{Department of Methodology and Statistics, Utrecht University}
  \city{Utrecht}
  \state{Utrecht}
  \country{The Netherlands}
}

\author{Anastasia Giachanou}
\orcid{0000-0002-7601-8667}
\email{a.giachanou@uu.nl}
\affiliation{%
  \institution{Department of Methodology and Statistics, Utrecht University}
  \city{Utrecht}
  \state{Utrecht}
  \country{The Netherlands}
}

\author{Daniel L. Oberski}
\orcid{0000-0001-7467-2297}
\email{d.l.oberski@uu.nl}
\affiliation{%
  \institution{Department of Methodology and Statistics, Utrecht University}
  \city{Utrecht}
  \state{Utrecht}
  \country{The Netherlands}
}

\author{Ayoub Bagheri}
\orcid{0000-0001-6366-2173}
\email{a.bagheri@uu.nl}
\affiliation{%
  \institution{Department of Methodology and Statistics, Utrecht University}
  \city{Utrecht}
  \state{Utrecht}
  \country{The Netherlands}
}




\begin{abstract}
{\bf Background:} 
    Generative models, especially large language models (LLMs), have shown remarkable progress in producing text that appears human-like. However, they often exhibit patterns that make their output easier to detect than text written by humans.
    
    {\bf Objectives:}
    In this paper, we investigate how explainable AI (XAI) methods can be used to reduce the detectability of AI-generated text  (AIGT) while also introducing a robust ensemble-based detection approach. 
    
    {\bf Methods:}
    We begin by training an ensemble classifier to distinguish AIGT from human-written text, then apply SHAP and LIME to identify tokens that most strongly influence its predictions. We propose four explainability-based token replacement strategies to modify these influential tokens. 
    
    {\bf Results:}
    Our findings show that these token replacement approaches can substantially diminish a single classifier's ability to detect AIGT. Human evaluators achieved only 47.0\% overall detection accuracy, showing that our strategies deceive both machines and humans. However, our ensemble classifier maintains strong performance across multiple languages and domains, showing that a multi-model approach can mitigate the impact of token-level manipulations.
    
    {\bf Conclusions:}
    Our findings show that XAI methods can make AIGT harder to detect by focusing on the most influential tokens. At the same time, they highlight the need for robust, ensemble-based detection strategies that can adapt to evolving approaches for hiding AIGT.
\end{abstract}




\maketitle


\section{Introduction}
\label{sec:intro}

Recent advancements in LLMs, such as the GPT series, have enabled the creation of highly fluent text that often resembles human writing~\cite{brown2020language}. Still, even the best models often show subtle signs of being artificial, like repeating phrases, using a limited range of words, having no typos, or showing a consistent writing style. These clues can help readers or automated tools detect that the text was written by AI~\cite{sadasivan2023can}. Being able to spot AIGT is important in various domains, such as in for protecting academic honesty, fighting misinformation, and keeping public trust. On the other hand, there are good reasons to make AI text less detectable, like tools that help people write in their own style~\cite{Yeh2024}, or AI summaries that aim to be especially clear and well-structured~\cite{nacke2024ai}.

This conflict between helpful and harmful uses of LLMs creates a challenge: how to keep their benefits while preventing misuse. One proposed solution is watermarking, which adds hidden patterns to generated text to make it more transparent~\cite{kirchenbauer2023watermarking}. However, recent research shows that these watermarks can be weakened or removed by changing the text, like paraphrasing~\cite{li2023towards}. Against this backdrop, our study asks:~\emph{Can explanation methods be used to both hide and detect AI-generated text?} We contribute in two ways. First, we develop an \textbf{ensemble‑based detector} that remains accurate across languages (English and Dutch) and domains (news, reviews, Twitter).  Second, we show how two explanation tools, SHapley Additive exPlanations (SHAP)~\cite{lundberg2017unified} and Local Interpretable Model-agnostic Explanations (LIME)~\cite{ribeiro2016should}, can be turned against a detector: we identify the most influential tokens and replace them via four targeted strategies (semantically similar human‑preferred words, part‑of‑speech‑constrained synonyms, GPT‑based substitutions, and GPT substitutions enriched with genre context).  We evaluate both \emph{detectability} (accuracy, F1) and \emph{textual fidelity} (BLEU, ROUGE), thereby quantifying the trade‑off between hiding AI signals and preserving meaning.  Our results show that token‑level manipulation can fool many single‑model detectors, yet a robust ensemble still performs strongly, underscoring the need to combine explainability, watermarking, and policy safeguards to manage the evolving landscape of AIGT.

The rest of this paper is organized as follows:
Section~\ref{sec:related} reviews related work on AI-generated text detection, watermarking, and adversarial rewriting.
Section~\ref{sec:dataset} describes the datasets and our data augmentation process.
In Section~\ref{sec:methodology}, we present our methodology, including details on model architectures, explainability approaches, and token replacement strategies.
Section~\ref{sec:experiments} reports experimental results, showing how different rewriting methods affect detection performance and textual fidelity.
Finally, Section~\ref{sec:discussion_conclusion} concludes the paper and discusses future research directions.

\section{Related Work}
\label{sec:related}

\citet{kirchenbauer2023watermarking} introduced an important \emph{red--green list} watermark. In this method, a secret hash divides the vocabulary at each token position into a preferred ``green'' group and a less preferred ``red'' group. The model then gives higher chances to the green tokens, so the generated text contains a hidden cryptographic signature. Light editing often keeps these watermarks, but bigger changes, like paraphrasing or replacing words with synonyms, can remove them~\citep{kirchenbauer2024reliability,li2023towards}. To make watermarks harder to notice, distortion-free methods hide information in the token-sampling \emph{order} instead of the probability distribution. However, these can also become weak if someone rewrites the text in a tricky way~\citep{christ2024undetectable}. Other approaches, called post-hoc schemes, add watermarks later by changing syntax templates or choosing context-aware synonyms~\citep{yang2023watermarking}, but they also have trouble surviving strong text changes.

Another important detection method takes advantage of the fact that language models (LMs) often choose very predictable words. \citet{gehrmann2019gltr} created a token-rank histogram tool to help people find words that are ``too likely'' in a sentence. Zero-shot methods like \emph{DetectGPT} look at the ``curvature'' of local log-probabilities; if the curvature is negative, it suggests the text was written by a machine~\citep{mitchell2023detectgpt}. \emph{Fast-DetectGPT} makes this process faster and more accurate for text from ChatGPT or GPT-4~\citep{bao2023fast}. Other detection methods use features such as normalized perturbed rank~\citep{su2023detectllm} and the amount of unused information in token probability distributions~\citep{zawistowski2024unused}. These ``white-box'' methods work well when they can directly access the model's probabilities, but they become less effective without this access. To solve this problem, ``DNA'' regeneration methods, like DNA-GPT~\citep{yang2023dna} and GPT Paternity Test~\citep{yu2023gpt}, do not need internal access; instead, they query a black-box model again and compare the new outputs for similarity.

Stylometric approaches instead focus on word choice, sentence structure, and how the text is organized and connected.~\citet{frohling2021feature} grouped 35 linguistic features into four categories, low diversity, repetition, incoherence, and lack of purpose, and trained SVM and random-forest models. More advanced work introduces entity-coherence graphs~\citep{liu2023coco} or journalism-specific style metrics~\citep{kumarage2023j}, which bolster detection on brief social-media texts. Though interpretable, these handcrafted methods may be narrower in scope and less transferable across domains.

A different direction uses fine-tuned neural networks to classify AI-generated text. OpenAI’s initial GPT-detector trained RoBERTa on mixed GPT-2 corpora, achieving roughly 95\% accuracy on in-domain data~\citep{solaiman2019release}. More recent detectors employ Longformer~\citep{beltagy2020longformer} or ELECTRA to handle longer contexts and detect texts from multiple generators~\citep{clark2020electra}. However, these systems often struggle with out-of-distribution domains. Recent ``cross-model'' or contrastive ensembles, such as Ghostbuster~\citep{verma2023ghostbuster}  and DeTeCtive~\citep{guo2024detective}, compare probability distributions from multiple open-source LMs to improve robustness.

Ensemble methods combine diverse signals to mitigate the weaknesses of individual detectors. Simple voting stacks of transformer-based classifiers outperform single models in multilingual tasks~\citep{nguyen2023stacking}. For example, recent work introduces ensemble-based approaches for detecting AI-generated text in educational content, achieving a favorable false-positive trade-off in academic scenarios~\citep{najjar2025detecting}. Our work follows this line by designing an explanation-oriented ensemble meant to withstand token-level adversarial edits.

Paraphrasing is one of the best ways to trick AI detectors. For example, \citet{krishna2023paraphrasing} showed that changing the wording can make DetectGPT miss almost all AI-written text. Rewriting the text many times makes it even harder to detect~\citep{sadasivan2023can}. Some new methods, like RADAR~\citep{hu2023radar} and OUTFOX~\citep{koike2024outfox}, are better at catching these changes because they are trained on paraphrased examples, but they still only recover some accuracy. Even small mistakes, like typos, can also confuse detectors~\citep{huang2024ai}. More advanced methods can handle some edits, but none can stop all tricks or rewrites.

Paraphrasing is seen as the strongest way to avoid AI text detectors:~\citet{krishna2023paraphrasing} found that rewriting text can reduce DetectGPT's ability to catch AI-generated text from 70\% to less than 5\%. Rewriting the text several times in a row (called "self-rewrites") makes it even harder to detect, while still keeping the original meaning~\citep{sadasivan2023can}. Defenses such as RADAR~\citep{hu2023radar} and OUTFOX~\citep{koike2024outfox} become stronger by training on paraphrased texts, and can recover 10–20 F\textsubscript{1} points. Recent studies also show that even simple changes at the letter level—like typos or swapping similar-looking letters—can confuse detectors and make them much less accurate~\citep{huang2024ai}. Beyond paraphrase-based evasion, reinforcement learning has emerged as a new paradigm for generating detector-resistant text. AuthorMist~\citep{david2025authormist} employs RL to iteratively refine text modifications that maximize evasion success while preserving semantic content, demonstrating that adaptive adversaries can systematically exploit detector weaknesses. Although using features like ensemble cross-probability or advanced watermarking can resist some small changes, no method is completely safe from strong rewriting attacks.

Explainability has become a cornerstone of modern NLP deployments~\citep{mohammadi2025explainability}. XAI tools like SHAP and LIME show which words or tokens have the biggest effect on a detector's decisions. For example, ~\citet{mitrovic2023chatgpt} used LIME to highlight important words and help people review short ChatGPT answers. But attackers can use this information in the opposite way, by changing or removing the most important words.~\citet{zhou2024humanizing} showed that using SHAP-based synonyms for key words can lower detector accuracy by 40–60\%. In our work, we carefully study this double-sided use of explanations, testing four types of word substitutions in two languages and different types of texts.

Beyond SHAP and LIME, alternative explainability approaches have been explored for NLP tasks. SyntaxShap~\citep{amara2024syntaxshap} extends SHAP by incorporating syntactic structure, assigning importance scores to phrase-level constituents rather than individual tokens, which can provide more linguistically meaningful explanations for text generation tasks. Similarly, Integrated Gradients has been applied to detect adversarial attacks on text classifiers~\citep{moraliyage2025explainable}, offering gradient-based attribution that traces predictions back through neural network layers. We chose SHAP and LIME for our study because they are model-agnostic and widely applicable to any classifier architecture, including our ensemble of transformers and XGBoost models. This flexibility allows practitioners to apply our token replacement framework regardless of the underlying detection model.

A natural question arises regarding the relationship between detection-based approaches and watermarking schemes. While watermarking proactively embeds identifiable patterns during text generation~\citep{kirchenbauer2023watermarking}, detection methods operate post-hoc on any text regardless of its origin. These approaches are complementary: watermarking provides provenance verification for known LLM outputs, while robust detection remains necessary for texts generated without watermarks or by adversaries who actively remove them~\citep{li2023towards}. Our XAI-based approach could potentially enhance watermarking schemes by identifying which tokens carry the strongest detection signal, informing where watermarks might be most resilient to modification. This synergy between explainability, detection, and watermarking represents a promising direction for comprehensive AI text governance.

Detectors often make mistakes with texts written by people using English as a second language.~\citet{liang2023gpt} found that seven popular GPT detectors wrongly marked more than 61\% of TOEFL essays by non-native English speakers as AI-generated, even though they were very accurate with essays by native speakers. Because of this bias, some teachers are moving away from strict rules and are now encouraging students to use AI tools as support, especially for those with different language backgrounds. Since LLMs can quickly create lots of convincing fake news, it is very hard for people alone to spot which content is real~\citep{zellers2019defending}. To improve detection in these situations, it helps to use a mix of classifiers and to add information about where the text comes from~\citep{fraser2025detecting}.

Recent use of AI chatbots for mental health support shows that people often think these systems are caring and human-like, which leads to real and meaningful conversations. But this can be risky, especially when users do not realize they are talking to an AI. ~\citet{song2024typing} found that people in severe distress sometimes seek help from chatbots and even see them as human. In the same way,~\citet{siddals2024happened} stressed the need for stronger safety rules and clearer information in AI mental health tools. These studies show how important it is to have good detection systems and to make it clear when someone is interacting with an AI, so that chatbots are used ethically in sensitive cases. Research also shows that no single method is perfect; reliable AI-generated text detection needs to use many signals, be tested against tricky cases, and give clear reasons for its decisions. These are the main goals of our work.

\section{Dataset}
\label{sec:dataset}
In this section, we detail the CLIN33 dataset~\cite{fivez2024clin33} used as our primary corpus and describe how it was augmented to expand its linguistic variability. We also present the portion of the AuTexTification dataset\footnote{\url{https://zenodo.org/records/10732813}} used to strengthen our ensemble models. The CLIN33 shared task corpus comprises human-written and AIGTs in English and Dutch, including three domains: news, tweets, and reviews. Each domain contains 200 samples from human authors and 200 samples from LLMs (GPT-4 and Vicu\~{n}a-13B), producing a total of 1,200 texts in English and 1,200 in Dutch. The human-written news samples come from well-known media outlets; the Dutch social media posts primarily discuss policy or health topics, whereas the English ones often revolve around social or sports events. Reviews typically cover literature in Dutch and various consumer goods in English. 

\begin{wraptable}{l}{0.65\textwidth}   
  \vspace{-1em}
  \centering
  \scriptsize
  \caption{\small Overview of data resources.}
  \label{tab:combined_data_overview}
  \begin{tabular}{p{2.5cm}p{3.4cm}p{3.0cm}}   
    \toprule
          & \textbf{AuTexTification (EN)} & \textbf{Augmented CLIN33} \\
    \midrule
    \textbf{Total Samples}   & 33{,}845 & 9{,}720 \\
    \textbf{Domains/Genres}  & Tweets, Legal, Wiki & News, Twitter, Reviews \\
    \textbf{Languages}       & English & English, Dutch \\
    \textbf{AI Models}       & BLOOM & GPT‑4 \\[-2pt]
                             & GPT‑3 & Vicu\~na‑13B \\
    \textbf{Human Sources}   & MultiEURLEX, Amazon, XSUM, TSATC, WikiLingua
                             & News, tweets, reviews \\
    \textbf{Class Ratio}     & 50.36\% / 49.64~\% & 50\% / 50\%\\
    \textbf{Augmentations}   & None & Yes (6 types, $\sim$3.5$\times$) \\
    \bottomrule
  \end{tabular}
  \vspace{-0.5em}
\end{wraptable}

Although CLIN33 offers balanced classes across two languages and three genres, its size remains relatively small for training large-scale models. To address this limitation and diversify the textual patterns, we created an augmented version of the dataset. We started with the original texts and applied a series of transformations to each sample. These transformations included synonym substitution, random word swaps, random insertions, targeted deletions, variations in spelling, and back-translation\footnote{Translating from English to Dutch and then back to English}. Each text thus produced several new variants that preserved the key semantics but introduced new surface forms. We generated an augmented pool of 9{,}720 instances by applying surface-level transformations (e.g., synonym substitution, swaps/insertions/deletions, spelling variants, and back-translation) to CLIN33 texts. From the 2,160 training texts (90\% of CLIN33), we applied transformations to create an average of 3.5 augmented variants per original text. Including the original texts, this yielded 9{,}720 total training instances (4.5 instances per source text on average). To avoid test leakage, augmentation was applied only to the training split (described below), while the held-out test set remained non-augmented.

As an additional resource for pre-training and fine-tuning, we employed a subset of the AuTexTification dataset\footnote{\url{https://sites.google.com/view/autextification}}, introduced by~\citet{sarvazyan2023overview}. While the complete dataset includes English and Spanish texts in five distinct domains, we selectively used the \emph{English-only} portion from tweets, legal, and wiki categories. This subset contains 33{,}845 texts designated for binary AI-detection (Task 1), split nearly evenly between human-written (50.36\%) and AI-generated (49.64\%) documents. The human content is drawn from sources such as MultiEURLEX, XSUM, Amazon Reviews, TSATC, and WikiLingua, while the AI-generated portion is produced by BLOOM (1B7, 3B, 7B1) and GPT-3 (babbage, curie, text-davinci-003). Because of its class balance and domain variety, this subset of AuTexTification serves as a valuable source of pre-training examples for our ensemble.
Table~\ref{tab:combined_data_overview} summarizes the essential attributes of the two datasets used to train and evaluate our methodology. The combined setup draws from the 33{,}845 samples in AuTexTification (English only) and the 9{,}720 augmented CLIN33 samples (both English and Dutch). We use AuTexTification Dataset for initial pre-training. We use 90\% of the original CLIN33 texts to construct training data (including augmentation), and reserve the remaining 10\% of original CLIN33 as an unchanged held-out test set.

We first created a stratified 90/10 split on the original (non-augmented) CLIN33 texts, preserving balance across language $\times$ domain $\times$ label. The held-out test set contains 240 original texts (120 human, 120 AI): 120 English and 120 Dutch, and 40 texts per domain per language (20 human, 20 AI). The remaining 90\% of the original CLIN33 texts were used for training/validation, and only that portion was augmented.

\begin{wrapfigure}{r}{0.48\textwidth}
  \vspace{-4.0em}
  \centering
  \footnotesize
  \setstretch{0.83}

  \captionof{algorithm}{Combined BERT Ensemble Structure}
  \label{alg:combinedmodel}

  \raggedright
  \textbf{Input:} Text $x$\\
  \textbf{Output:} Predicted label \{\emph{human}, \emph{AI}\}\\[2pt]

  \begin{algorithmic}[1]
    \Function{PretrainEnsemble}{$x$}
      \State $\text{bert\_out}   \leftarrow \text{BERTModel}_{\text{frozen}}(x)$
      \State $\text{xlmr\_out}   \leftarrow \text{XLMRobertaModel}_{\text{frozen}}(x)$
      \State $\text{distil\_out} \leftarrow \text{DistilBERTModel}_{\text{frozen}}(x)$
      \State \Return $(\text{bert\_out}, \text{xlmr\_out}, \text{distil\_out})$
    \EndFunction

    \Function{FineTuneEnsemble}{$x$}
      \State $\text{bert\_out\_fresh}   \leftarrow \text{BERTModel}_{\text{fresh}}(x)$
      \State $\text{xlmr\_out\_fresh}   \leftarrow \text{XLMRobertaModel}_{\text{fresh}}(x)$
      \State $\text{distil\_out\_fresh} \leftarrow \text{DistilBERTModel}_{\text{fresh}}(x)$
      \State \Return $(\text{bert\_out\_fresh}, \text{xlmr\_out\_fresh}, \text{distil\_out\_fresh})$
    \EndFunction

    \Function{CombinedModel}{$x$}
      \State $(\text{froz\_bert}, \text{froz\_xlmr}, \text{froz\_distil})
        \leftarrow \Call{PretrainEnsemble}{x}$
      \State $(\text{fresh\_bert}, \text{fresh\_xlmr}, \text{fresh\_distil})
        \leftarrow \Call{FineTuneEnsemble}{x}$

      \State $\text{concat\_outs} \leftarrow \text{concatenate}[\text{froz\_bert}, \text{froz\_xlmr}, \text{froz\_distil},$
      \Statex\hspace{\algorithmicindent}
        $\text{fresh\_bert}, \text{fresh\_xlmr}, \text{fresh\_distil}]$

      \State $\text{dense\_out} \leftarrow \text{DenseLayer}(\text{concat\_outs})$
      \State $\hat{y} \leftarrow \sigma(\text{dense\_out})$
      \If{$\hat{y} \geq 0.5$}
        \State \Return \emph{AI}
      \Else
        \State \Return \emph{human}
      \EndIf
    \EndFunction
  \end{algorithmic}
\end{wrapfigure}

\section{Methodology}
\label{sec:methodology}
We used an approach that integrates several models for text classification to address the AIGT detection task. For XGBoost we apply lowercase/URL removal/punctuation removal/lemmatization. For transformer models we use the native tokenizer with no extra preprocessing. We fine-tune bert-base-multilingual-uncased, distilbert-base-multilingual-cased, and xlm-roberta-base for both English and Dutch.
\paragraph{\textbf{Baselines Models.}} We built an XGBoost classifier by first transforming the text into numerical features with a TF-IDF vectorizer set to a maximum of 5000 features. The XGBoost model was configured with use\_label\_encoder=False and eval\_metric='logloss'. Training used the TF-IDF representations from the training set, and final predictions on the test set were evaluated using precision, recall, F1, and accuracy. Also, We fine-tuned BERT-base (bert-base-multilingual-uncased), DistilBERT (distilbert-base-multilingual-cased), and  XLM-RoBERTa (xlm-roberta-base). The best checkpoint was chosen based on the highest F1 score on a validation subset. We then evaluated the final models on the test data.

\paragraph{\textbf{Ensemble Model.}}
We built an ensemble of multiple transformers instead of relying on a single one to classify human and AIGT), as inspired by~\citet{mohammadi2023towards}. In this method, we use the strengths of each model and create an ensemble that can be fine-tuned on similar datasets. This helps the model adapt to different domains and reduces problems caused by limited data, as the features learned by each part complement each other. Specifically, we integrated bert-base-multilingual-uncased, distilbert-base-multilingual-cased, and xlm-roberta-base. Each transformer produced a representation of the input text, and we concatenated these representations before passing them through a dense layer with L2 regularization for final classification.

\begin{figure*}[!ht]
\centering
\includegraphics[width=0.7\linewidth]{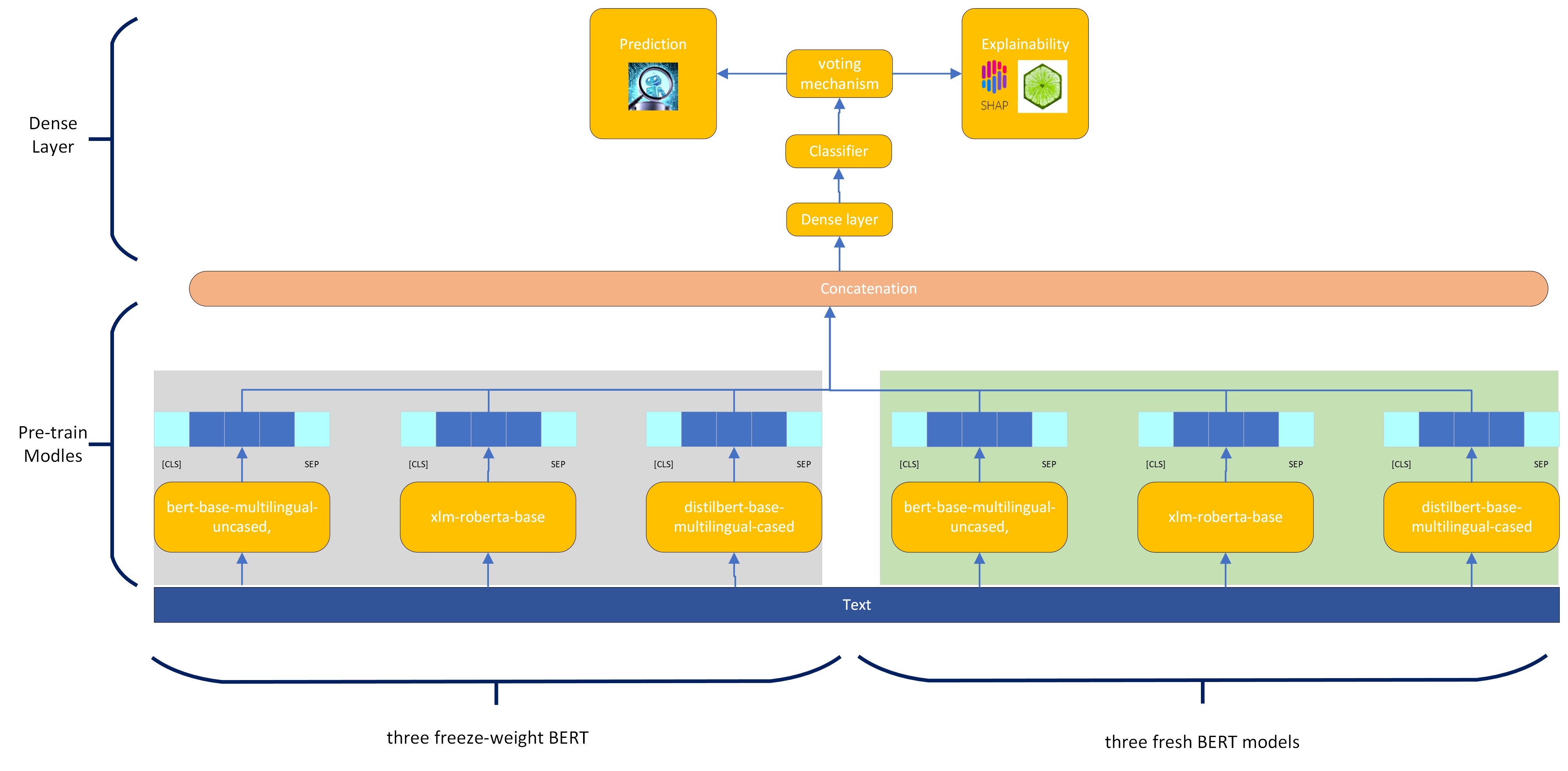}
\vspace{-5pt}
\caption{\small The architecture of the proposed ensemble model. Outputs of three frozen BERT models (left) and three fresh BERT models (right) are concatenated and passed through a dense layer followed by a sigmoid activation (and threshold at 0.5) for final binary predictions.}
\label{fig:ms-structure}
\end{figure*}

We first built a \emph{Combined BERT} model by training three BERT models on the AuTexTification training set. We optimized the network's hyperparameters and evaluated performance on the AuTexTification test set, focusing on F1 and average metrics across folds. After training, we froze these model weights to retain the features they had learned and reduce the risk of overfitting. We next created a \emph{combined model} that merges three BERT models with frozen weights (trained on AuTexTification) and three fresh BERT models with unfrozen parameters. These fresh models were fine-tuned on our newly augmented dataset Such an ensemble synergy aligns with~\citet{fraser2025detecting}, who highlight the benefits of multi-model integration in detecting AI-generated text. Algorithm~\ref{alg:combinedmodel} outlines the training and process for our combined BERT ensemble. During training, we minimize the binary cross-entropy loss on each branch using the augmented data. We then unify the outputs in a final fully connected layer with L2 regularization. Figure~\ref{fig:ms-structure} shows the architecture, with frozen models on the left (gray) and fresh models on the right (green).

\subsection{Experimental Setup}
\label{subsec:setup}

We trained our models using the Hugging Face Transformers library\footnote{\url{https://huggingface.co/}} and set a maximum token length of 512. Hyperparameter tuning was done through a random search over learning rates (ranging from \(1\times 10^{-5}\) to \(1\times 10^{-4}\)) and batch sizes of 32, 64, or 128. We used a cosine decay schedule for learning rate adjustments, with 200 warm-up steps and an early-stopping patience of 5 epochs based on validation loss. 

We trained on the augmented training split and selected hyperparameters using a validation split drawn from training data. Final evaluation was performed on a held-out non-augmented CLIN33 test set of 240 texts (120 human, 120 AI), stratified across two languages and three domains (40 texts per domain per language). This same fixed test set was used for all experiments to ensure fair comparisons. The hyperparameters and early-stopping rules were adjusted using the validation split. After training, we evaluated the system on the remaining data, measuring precision, recall, F1-score, and accuracy.  By combining different transformers, freezing selectively learned features, and training fresh models in parallel, this ensemble method balanced the advantages of robust pre-trained representations with the adaptability needed for the new dataset.

\paragraph{Computational Resources.} Table~\ref{tab:computational_cost} summarizes the computational requirements of each model in our experiments. All training was conducted on NVIDIA A100 GPUs (40GB memory). The ensemble model, while requiring substantially more resources than individual models, remains tractable for research settings and can be parallelized across multiple GPUs to reduce wall-clock training time. For inference, the ensemble processes samples sequentially through each component, with the total inference time being the sum of individual model times plus the final dense layer computation. In production scenarios, inference latency could be reduced through model distillation or parallel execution of the ensemble components.

\begin{table}[ht]
\centering
\small
\caption{Computational cost comparison of detection models. Training times are approximate and based on the CLIN33 dataset. Inference times represent average per-sample latency on GPU. Frozen branch $\approx$454M parameters; fresh branch $\approx$454M parameters; total parameters loaded for the ensemble $\approx$908M (+ classifier head).}
\label{tab:computational_cost}
\begin{tabular}{lcccc}
\toprule
\textbf{Model} & \textbf{Parameters} & \textbf{Training Time} & \textbf{Inference (ms)} & \textbf{GPU Memory} \\
\midrule
XGBoost & $\sim$50K & $\sim$5 min & $\sim$1 & CPU only \\
BERT-base & 110M & $\sim$2 h & $\sim$50 & $\sim$4 GB \\
DistilBERT & 66M & $\sim$1.5 h & $\sim$30 & $\sim$2 GB \\
XLM-RoBERTa & 278M & $\sim$3 h & $\sim$60 & $\sim$6 GB \\
\midrule
Ensemble (ours) & $\sim$908M & $\sim$8 h total & $\sim$200 & $\sim$16 GB \\
\bottomrule
\end{tabular}
\end{table}

\subsection{Explainability-driven detection}
\label{sec:explain}

In this section, we identify the words that most strongly influence an \emph{AI-generated} prediction. Our approach uses two instance-level explanation tools: SHAP and LIME. Although they both operate on individual examples, SHAP uses game-theoretic principles to estimate a token’s overall (global) influence, while LIME explores small local perturbations to approximate importance. We take the absolute values of these importance scores $\phi_j$ (for SHAP) or coefficients (for LIME) to capture both positive and negative contributions.

  \setlength\intextsep{0.3em}   
  \setlength\columnsep{1em}     
  \begin{wrapfigure}[5]{r}{0.40\textwidth}  
    \vspace*{-0.2em}                        
    \begin{minipage}{\linewidth}\centering\scriptsize
      \[
        \phi_j
          = \sum_{S \subseteq \{1,\dots,m\}\setminus\{j\}}
            \frac{|S|!\,\bigl(m-|S|-1\bigr)!}{m!}\,
            \bigl[f(S \cup \{j\}) - f(S)\bigr].
      \]
    \end{minipage}
    \vspace*{-0.4em}                        
  \end{wrapfigure}

We follow the methodology in~\cite{mohammadi2024transparent}, in this framework, we only analyze sentences that the model \emph{correctly} classifies, ensuring the tokens we highlight have genuinely contributed to accurate decisions. We then average each token’s importance within the sentence to get a consistent measure of its overall effect. This step helps us confirm that the extracted terms are meaningful indicators of the model’s reasoning. Given a trained model \(f\) and an instance \(x\) composed of tokens \((t_1, t_2, \ldots, t_m)\), SHAP assigns a Shapley value \(\phi_j\) to each token \(t_j\). These values are computed by measuring how much including or excluding \(t_j\) changes the model’s output, across all possible subsets of tokens. Formally:

Here, \(f(S)\) is the model's prediction when only the tokens in set \(S\) are present. The weighting term ensures each token's effect is fairly distributed among all subsets~\cite{lundberg2017unified}. LIME, on the other hand, makes small changes near the original text to learn a simpler, local model that approximates \(f\). It then assigns coefficients reflecting each token’s importance for that instance~\cite{ribeiro2016should}. Although LIME focuses more on local neighborhoods, it also provides per-token contributions based on how the model’s prediction shifts when individual tokens are perturbed. We also consider a random token selection. This lets us compare how much improvement we gain by targeting the tokens that SHAP or LIME indicates are most influential. By using both a global (SHAP) and local (LIME) perspective, we can better trust that the identified tokens genuinely drive the model's decision. In our subsequent steps, we use these tokens to guide targeted replacements and measure how effectively such edits can mask AI-generated content.

\subsection{Token Replacement Strategies}
Once we identify influential tokens, we modify them using several approaches designed to make the text appear more human-like. We propose four rewriting strategies summarized in Table~\ref{tab:token_replacement}.

\begin{table}[ht]
\centering
\scriptsize
\renewcommand{\arraystretch}{1.2}
\caption{Summary of token replacement strategies}
\label{tab:token_replacement}
\begin{tabular}{p{4.8cm}p{9.2cm}}
\toprule
\textbf{Strategy} & \textbf{Description} \\
\midrule
\textbf{Human Similar Word Replacement (HSR)} 
 & A Word2Vec model is trained on all human-labeled texts in the training set to learn an embedding space that captures typical human usage. \\
\textbf{Part-of-Speech Replacement (PSR)} 
 & HSR is refined by ensuring that part-of-speech tags match. \\
\textbf{GPT-based Replacement (GPT)} 
 & A generative model (GPT-4o-mini) is queried using the prompt: 
   \texttt{"Replace '\{token\}' with a more human-like word in this text: '\{text\}'"}. \\
\textbf{GPT-based Replacement with Genre Context (GPT+Genre)} 
 & The GPT-based method is extended by including explicit genre information (e.g., \emph{tweets} or \emph{reviews}) in the prompt:
   \texttt{"Replace '\{token\}' in this \{domain\} text with a more human-like word: '\{text\}'"}. \\
\bottomrule
\end{tabular}
\end{table}

In \emph{Strategy HSR}, each influential token is replaced with its nearest neighbor in the Word2Vec space, provided that the neighbor is frequently observed in human texts. In \emph{Strategy PSR}, if a token \(t\) is tagged as a noun, for instance, a Word2Vec neighbor that is also tagged as a noun is selected, preserving grammatical structure. In \emph{Strategy GPT}, a replacement that appears more \emph{human-like} is proposed based on the surrounding sentence. While the suggestions are context-sensitive, certain LLM-specific patterns may still be retained if the generative model reintroduces specific phrasing. In \emph{Strategy GPT+Genre}, incorporating domain-specific information allows the generative model to propose replacements that align more closely with the original style. However, this added context may also amplify distinctive AI traits.  Finally, after substituting the most influential tokens in each text, we assess performance along two dimensions: (1) changes in detectability and (2) fidelity to the original content.

\subsection{Evaluation Metrics}
\label{subsec:evaluation_metrics}

We use a combination of detection and text-similarity metrics to evaluate how rewriting strategies affect our models' ability to detect AIGT, as well as how much the rewritten text deviates from its original form. First, once we rewrite an AIGT via HSR, PSR, GPT, or GPT+Genre. Next, we investigate how different rewriting scenarios affect AI-text detection. We adopt the adversarial rewriting perspective discussed in \citet{fraser2025detecting}, 
where paraphrased or partially edited texts can undermine naive detectors. We run it through each classifier to check whether the label remains \emph{AI} or flips to \emph{human}. We then calculate precision, recall, F1, and accuracy to capture each model’s classification performance to detect AI/human texts under these modified inputs. Additionally, we compute the \emph{flip rate}, defined as the fraction of AI texts whose predicted label changes to \emph{human} after rewriting; a higher flip rate implies that the rewriting method is more effective in evading detection. In parallel, we calculate the degree of textual change introduced by each rewriting approach. We use BLEU, ROUGE-1, and ROUGE-L to measure the overlap between the original text and its rewritten form, with higher scores showing fewer modifications. By combining detection metrics, and text-similarity scores, we gain a clear view of how effectively each rewriting method hides AI-generated content and how much it changes the original text. 

\section{Experimental Analysis and Results}
\label{sec:experiments}

In this section, we compare five classification models (XGBoost, BERT-base, DistilBERT, XLM-RoBERTa, and our \textbf{Ensemble} approach) across various token-replacement scenarios for the detection of AIGT, Consistent with~\citet{fraser2025detecting}, we anticipate that multi-domain evaluation is critical for robust results. Our main objective is to observe how different rewriting methods (HSR, PSR, GPT, GPT+Genre) combined with token-selection strategies (SHAP, LIME, Random) affect each model’s ability to detect generated text.

Table~\ref{tab:lang_domain_updated} shows each model's performance on \emph{unmodified} AIGT across two languages (English, Dutch) and three domains (news, reviews, and X/Twitter). As can be seen, although certain single models perform well on specific subsets (e.g., DistilBERT in English news, XGBoost in English Twitter), the \textbf{Ensemble} approach tends to achieve the best performance in \emph{most} cases. This pattern is particularly clear in Dutch news, where the Ensemble model achieves near-perfect scores (F1=0.99, Precision=0.98, Recall=1.00). Even when other models have high recall or precision on certain tasks, they often fall behind on at least one metric (e.g., XLM-RoBERTa predicts too many positives, increasing recall but lowering precision). showing strong cross-language adaptability. This result agrees with ~\citet{fraser2025detecting}, who emphasize the importance of multi-lingual calibration for robust detection.

\begin{table}[ht]
\centering
\scriptsize
\renewcommand{\arraystretch}{1.2}
\caption{Model performance by language and domain on the original (unmodified) test set. \textbf{Bold} indicates the best result within each language and domain block.}
\label{tab:lang_domain_updated}
\begin{tabular}{llcccc|llcccc}
\toprule
\multicolumn{6}{c}{\textbf{English Dataset}} & \multicolumn{6}{c}{\textbf{Dutch Dataset}} \\
\cmidrule(lr){1-6} \cmidrule(lr){7-12}
\textbf{Model} & \textbf{Domain} & \textbf{Acc} & \textbf{Prec} & \textbf{Rec} & \textbf{F1} 
               & \textbf{Model} & \textbf{Domain} & \textbf{Acc} & \textbf{Prec} & \textbf{Rec} & \textbf{F1} \\
\midrule
\textbf{--- News ---} &       &     &     &     &     
& \textbf{--- News ---} &       &     &     &     &     \\
XGBoost     
  & news    & 0.89 & 0.93 & 0.84 & 0.88 
  & XGBoost     
  & news    & 0.85 & 0.88 & 0.88 & 0.88 \\
BERT-base  
  & news    & 0.94 & 0.97 & 0.91 & 0.94
  & BERT-base  
  & news    & 0.68 & 0.79 & 0.56 & 0.66 \\
DistilBERT 
  & news    & 0.92 & 0.90 & \textbf{0.95} & 0.92
  & DistilBERT 
  & news    & 0.68 & 0.73 & 0.66 & 0.69 \\
XLM-RoBERTa
  & news    & 0.49 & 0.49 & 1.00 & 0.66
  & XLM-RoBERTa
  & news    & 0.53 & 0.53 & \textbf{1.00} & 0.69 \\
Ensemble   
  & news    & \textbf{0.95} & \textbf{1.00} & 0.90 & \textbf{0.95}
  & Ensemble   
  & news    & \textbf{0.99} & \textbf{0.98} & \textbf{1.00} & \textbf{0.99} \\
\midrule
\textbf{--- Reviews ---} &    &     &     &     &     
& \textbf{--- Reviews ---} &    &     &     &     &     \\
XGBoost    
  & reviews & 0.51 & 0.57 & 0.59 & 0.58
  & XGBoost    
  & reviews & 0.53 & 0.51 & 0.46 & 0.48 \\
BERT-base  
  & reviews & 0.50 & 0.54 & 0.62 & 0.58
  & BERT-base  
  & reviews & 0.40 & 0.35 & 0.30 & 0.32 \\
DistilBERT 
  & reviews & 0.65 & 0.68 & 0.71 & 0.69
  & DistilBERT 
  & reviews & 0.42 & 0.38 & 0.50 & 0.43 \\
XLM-RoBERTa
  & reviews & 0.56 & 0.56 & \textbf{0.90} & 0.69
  & XLM-RoBERTa
  & reviews & 0.43 & 0.43 & \textbf{0.60} & 0.48 \\
Ensemble   
  & reviews & \textbf{0.67} & \textbf{0.71} & 0.68 & \textbf{0.70}
  & Ensemble   
  & reviews & \textbf{0.60} & \textbf{0.53} & 0.44 & \textbf{0.49} \\
\midrule
\textbf{--- Twitter / X ---} &   &     &     &     &     
& \textbf{--- Twitter / X ---} &   &     &     &     &     \\
XGBoost    
  & twitter & 0.91 & \textbf{0.92} & 0.98 & 0.95
  & XGBoost    
  & twitter & 0.80 & 0.85 & 0.78 & 0.81 \\
BERT-base  
  & twitter & 0.90 & 0.86 & 0.98 & 0.92
  & BERT-base  
  & twitter & 0.80 & 0.85 & 0.78 & 0.81 \\
DistilBERT 
  & twitter & 0.90 & 0.86 & 0.98 & 0.92
  & DistilBERT 
  & twitter & 0.84 & 0.86 & 0.80 & 0.83 \\
XLM-RoBERTa
  & twitter & 0.60 & 0.53 & \textbf{1.00} & 0.69
  & XLM-RoBERTa
  & twitter & 0.62 & 0.60 & \textbf{0.90} & 0.72 \\
Ensemble
  & twitter & \textbf{0.93} & 0.86 & \textbf{1.00} & \textbf{0.93}
  & Ensemble
  & twitter & \textbf{0.87} & \textbf{0.90} & 0.85 & \textbf{0.87} \\
\midrule
\multicolumn{6}{c}{\textbf{English Average}} & \multicolumn{6}{c}{\textbf{Dutch Average}} \\
\cmidrule(lr){1-6} \cmidrule(lr){7-12}
XGBoost    & avg & 0.77 & 0.81 & 0.80 & 0.80
  & XGBoost    & avg & 0.73 & 0.75 & 0.71 & 0.72 \\
BERT-base  & avg & 0.78 & 0.79 & 0.84 & 0.81
  & BERT-base  & avg & 0.63 & 0.66 & 0.55 & 0.60 \\
DistilBERT & avg & 0.82 & 0.81 & 0.88 & 0.84
  & DistilBERT & avg & 0.65 & 0.66 & 0.65 & 0.65 \\
XLM-RoBERTa& avg & 0.55 & 0.53 & 0.97 & 0.68
  & XLM-RoBERTa& avg & 0.53 & 0.52 & 0.83 & 0.63 \\
Ensemble   & avg & \textbf{0.85} & \textbf{0.86} & 0.86 & \textbf{0.86}
  & Ensemble   & avg & \textbf{0.82} & \textbf{0.80} & 0.76 & \textbf{0.78} \\
\bottomrule
\end{tabular}
\end{table}

We next investigate how different rewriting scenarios affect AI-text detection. In particular, we apply four rewriting strategies and combine each of these with one of three token-selection strategies: SHAP (prioritizes globally influential tokens), LIME (locally influential tokens), or Random (tokens chosen arbitrarily). Table~\ref{tab:all_models_rewriting_f1acc} presents performance metrics for each model under all these rewriting conditions.

\begin{table*}[ht]
\centering
\caption{\small F1 score and Accuracy of each detector under every rewriting scenario.
The best metric in a column is \textbf{bolded}.
All results are the mean over 4 runs (different random seeds).
\footnotesize{$\Delta$ (pp) $=(\text{Scenario Value}-\text{Original Value})\times100$. Note: $\Delta$ represents absolute percentage point change, not relative percent change.}}
\label{tab:all_models_rewriting_f1acc}
\renewcommand{\arraystretch}{1.4}
\resizebox{\textwidth}{!}{
\begin{tabular}{l l
c c  c c  c c  c c  c c}
\toprule
\multirow{2}{*}{\textbf{Explain}} & \multirow{2}{*}{\textbf{Strategy}}
& \multicolumn{2}{c}{\textbf{XGBoost}}
& \multicolumn{2}{c}{\textbf{BERT-base}}
& \multicolumn{2}{c}{\textbf{DistilBERT}}
& \multicolumn{2}{c}{\textbf{XLM-RoBERTa}}
& \multicolumn{2}{c}{\textbf{Ensemble}}\\
\cmidrule(lr){3-4}\cmidrule(lr){5-6}\cmidrule(lr){7-8}\cmidrule(lr){9-10}\cmidrule(lr){11-12}
& & \textbf{F1 ($\Delta$ pp)} & \textbf{Acc ($\Delta$ pp)}
  & \textbf{F1 ($\Delta$ pp)} & \textbf{Acc ($\Delta$ pp)}
  & \textbf{F1 ($\Delta$ pp)} & \textbf{Acc ($\Delta$ pp)}
  & \textbf{F1 ($\Delta$ pp)} & \textbf{Acc ($\Delta$ pp)}
  & \textbf{F1 ($\Delta$ pp)} & \textbf{Acc ($\Delta$ pp)}\\
\midrule
\emph{NoRewriting} & Original
& 0.81 & 0.80
& 0.74 & 0.75
& 0.79 & 0.77
& 0.68 & 0.54
& \textbf{0.83} & \textbf{0.83}\\
\midrule
\textbf{SHAP} & HSR
& 0.73 (\largest{-8 pp}) & 0.75 (\largest{-5 pp})
& 0.71 (-3 pp) & 0.72 (-3 pp)
& 0.76 (-3 pp) & 0.75 (-2 pp)
& 0.68 (\smallest{0 pp}) & 0.54 (\smallest{0 pp})
& \textbf{0.78} (-5 pp) & \textbf{0.80} (-3 pp)\\
\textbf{SHAP} & PSR
& 0.78 (\largest{-3 pp}) & 0.78 (\largest{-2 pp})
& 0.73 (-1 pp) & 0.74 (-1 pp)
& 0.78 (-1 pp) & 0.77 (0 pp)
& 0.68 (\smallest{0 pp}) & 0.54 (\smallest{0 pp})
& \textbf{0.82} (-1 pp) & \textbf{0.82} (-1 pp)\\
\textbf{SHAP} & GPT
& 0.79 (-2 pp) & 0.79 (-1 pp)
& 0.73 (-1 pp) & 0.74 (-1 pp)
& 0.77 (-2 pp) & 0.76 (-1 pp)
& 0.68 (\smallest{0 pp}) & 0.55 (\smallest{+1 pp})
& \textbf{0.80} (\largest{-3 pp}) & \textbf{0.81} (\largest{-2 pp})\\
\textbf{SHAP} & GPT+Genre
& \textbf{0.80} (-1 pp) & \textbf{0.80} (0 pp)
& 0.73 (-1 pp) & 0.74 (-1 pp)
& 0.77 (-2 pp) & 0.76 (-1 pp)
& 0.68 (\smallest{0 pp}) & 0.55 (\smallest{+1 pp})
& 0.79 (\largest{-4 pp}) & 0.80 (\largest{-3 pp})\\
\midrule
\textbf{LIME} & HSR
& 0.25 (\largest{-56 pp}) & 0.48 (\largest{-32 pp})
& 0.56 (-18 pp) & 0.62 (-13 pp)
& 0.62 (-17 pp) & 0.65 (-12 pp)
& \textbf{0.67} (\smallest{-1 pp}) & 0.52 (\smallest{-2 pp})
& 0.64 (-19 pp) & \textbf{0.69} (-14 pp)\\
\textbf{LIME} & PSR
& 0.56 (\largest{-25 pp}) & 0.63 (\largest{-17 pp})
& 0.64 (-10 pp) & 0.67 (-8 pp)
& 0.71 (-8 pp) & 0.71 (-6 pp)
& 0.66 (\smallest{-2 pp}) & 0.52 (\smallest{-2 pp})
& \textbf{0.73} (-10 pp) & \textbf{0.75} (-8 pp)\\
\textbf{LIME} & GPT
& 0.57 (\largest{-24 pp}) & 0.64 (\largest{-16 pp})
& 0.73 (-1 pp) & 0.74 (-1 pp)
& 0.78 (-1 pp) & 0.77 (0 pp)
& 0.68 (\smallest{0 pp}) & 0.55 (\smallest{+1 pp})
& \textbf{0.83} (\smallest{0 pp}) & \textbf{0.83} (0 pp)\\
\textbf{LIME} & GPT+Genre
& 0.60 (\largest{-21 pp}) & 0.65 (\largest{-15 pp})
& \textbf{0.75} (\smallest{+1 pp}) & \textbf{0.75} (0 pp)
& 0.77 (-2 pp) & 0.76 (-1 pp)
& 0.68 (0 pp) & 0.55 (\smallest{+1 pp})
& \textbf{0.81} (-2 pp) & \textbf{0.81} (-2 pp)\\
\midrule
\textbf{Random} & HSR
& 0.73 (-8 pp) & 0.74 (-6 pp)
& 0.63 (\largest{-11 pp}) & 0.67 (\largest{-8 pp})
& 0.68 (\largest{-11 pp}) & 0.69 (\largest{-8 pp})
& 0.67 (\smallest{-1 pp}) & 0.53 (\smallest{-1 pp})
& \textbf{0.73} (-10 pp) & \textbf{0.75} (\largest{-8 pp})\\
\textbf{Random} & PSR
& 0.76 (-5 pp) & 0.76 (-4 pp)
& 0.65 (\largest{-9 pp}) & 0.68 (\largest{-7 pp})
& 0.74 (-5 pp) & 0.73 (-4 pp)
& 0.67 (\smallest{-1 pp}) & 0.53 (\smallest{-1 pp})
& \textbf{0.78} (-5 pp) & \textbf{0.79} (-4 pp)\\
\textbf{Random} & GPT
& 0.74 (\largest{-7 pp}) & 0.75 (\largest{-5 pp})
& 0.74 (0 pp) & 0.75 (0 pp)
& 0.77 (-2 pp) & 0.76 (-1 pp)
& 0.68 (\smallest{0 pp}) & 0.55 (\smallest{+1 pp})
& \textbf{0.81} (-2 pp) & \textbf{0.82} (-1 pp)\\
\textbf{Random} & GPT+Genre
& 0.75 (\largest{-6 pp}) & 0.75 (\largest{-5 pp})
& 0.73 (-1 pp) & 0.74 (-1 pp)
& \textbf{0.78} (-1 pp) & 0.77 (0 pp)
& 0.68 (0 pp) & 0.55 (\smallest{+1 pp})
& \textbf{0.81} (-2 pp) & \textbf{0.81} (-2 pp)\\
\bottomrule
\end{tabular}}
\end{table*}

\paragraph{\textbf{General Trends.}}
When no rewriting is applied (\emph{NoRewriting, Original}), most models exhibit reasonably strong performance (precision and recall around 0.75--0.83), with the Ensemble reaching the highest overall F1 (0.83) and accuracy (0.83). After rewriting, simpler models often detect less, especially if the replaced tokens were the ones they used to identify AI patterns.

\paragraph{\textbf{Comparison of Token-Selection Approaches.}}
SHAP or LIME-based replacements often make larger drops in detection metrics than Random replacements, since replacing highly influential tokens more effectively removes AI indicators. For instance, comparing LIME+HSR vs.\ Random+HSR, XGBoost’s F1 can decrease from around 0.73 to 0.25 if LIME is used, indicating a big loss in detection ability once key tokens are perturbed. On the other hand, random token replacement leaves more of the AI signals intact, so detection remains higher.

\paragraph{\textbf{Influence of Rewriting Methods.}}
HSR and PSR replace tokens with synonyms from a human corpus, often preserving the original text’s structure. This is reflected in relatively high BLEU scores (e.g., over 90\% in the SHAP+HSR or SHAP+PSR rows), meaning minimal rewriting. Nonetheless, these small edits can still mislead detectors if the replaced tokens were crucial. GPT-based rewrites typically lower BLEU scores to the range of 0.76–0.86, reflecting more substantial paraphrases. In turn, many detectors (e.g., XGBoost or XLM-RoBERTa) experience an increase in recall but a decrease in precision, or the opposite, because the text shifts away from familiar AI phrasing. GPT+Genre further contextualizes the rewrite with domain hints (e.g.\ write it like a news article), causing domain-sensitive lexical changes. This can introduce new phrasing that confuses detectors reliant on domain-specific cues.

\paragraph{\textbf{Model-Specific Observations.}}
XLM-RoBERTa often shows high recall (sometimes hitting 1.0), but its precision remains around 0.52, leading to only modest changes in F1. DistilBERT is more balanced yet remains vulnerable when its top keywords are replaced by synonyms or paraphrases. BERT-base typically holds a moderate advantage across scenarios but can be undermined in LIME-based rewrites (e.g., LIME+HSR, where BERT-base F1 falls to about 0.56). By contrast, the \textbf{Ensemble} approach consistently retains the highest or near-highest F1 and accuracy. For example, under LIME+GPT, the Ensemble still attains an F1 of 0.83, substantially above XGBoost's 0.57 or BERT-base's 0.73. Similarly, in SHAP plus GPT+Genre, the Ensemble preserves a strong F1 of about 0.79, while other models vary more dramatically. Also, We show text similarity scores (BLEU, ROUGE1, ROUGEL) to capture how extensively each rewrite modifies the original text.

\paragraph{\textbf{Text Similarity Implications.}}
 Table~\ref{tab:rewriting_similarity} presents text similarity scores (BLEU, ROUGE1, ROUGEL) under all these rewriting conditions. Looking at BLEU and ROUGE scores confirms that GPT-based methods, especially GPT+Genre, introduce the largest textual shifts. Replacements guided by SHAP+HSR or SHAP+PSR keep BLEU at or above 0.90 (0.93 and 0.97 respectively), while LIME/Random-based HSR achieve lower BLEU scores (~0.84). GPT-based rewrites show BLEU scores ranging from 0.76 to 0.86 depending on the explainability method used. This trade-off highlights how more extensive rewriting can more thoroughly remove AI signatures at the expense of text fidelity.

\begin{wraptable}{r}{0.62\textwidth}   
  \vspace{-0.8em}                      
  \centering
  \scriptsize
  \caption{\small BLEU, ROUGE‑1 and ROUGE‑L scores for every rewriting scenario.}
  \label{tab:rewriting_similarity}
  \renewcommand{\arraystretch}{1.2}
  \begin{tabular}{l l c c c}
    \toprule
    \textbf{Explain} & \textbf{Strategy} & \textbf{BLEU ($\Delta$)} & \textbf{ROUGE‑1 ($\Delta$)} & \textbf{ROUGE‑L ($\Delta$)}\\
    \midrule
    \emph{NoRewriting} & Original & 1.00 & 1.00 & 1.00\\
    \midrule
    \textbf{SHAP} & HSR & 0.93 (-7 pp) & 0.97 (-3 pp) & 0.97 (-3 pp)\\
    \textbf{SHAP} & PSR & \textbf{0.97} (\smallest{-3 pp}) & \textbf{0.99} (\smallest{-1 pp}) & \textbf{0.99} (\smallest{-1 pp})\\
    \textbf{SHAP} & GPT & 0.86 (-14 pp) & 0.93 (-7 pp) & 0.93 (-7 pp)\\
    \textbf{SHAP} & GPT+Genre & 0.86 (-14 pp) & 0.94 (-6 pp) & 0.94 (-6 pp)\\
    \midrule
    \textbf{LIME} & HSR & 0.84 (-16 pp) & 0.92 (-8 pp) & 0.92 (-8 pp)\\
    \textbf{LIME} & PSR & 0.90 (-10 pp) & 0.96 (-5 pp) & 0.95 (-5 pp)\\
    \textbf{LIME} & GPT & 0.76 (\largest{-24 pp}) & 0.84 (\largest{-16 pp}) & 0.84 (\largest{-16 pp})\\
    \textbf{LIME} & GPT+Genre & 0.76 (\largest{-24 pp}) & 0.85 (-15 pp) & 0.84 (\largest{-16 pp})\\
    \midrule
    \textbf{Random} & HSR & 0.84 (-16 pp) & 0.92 (-8 pp) & 0.92 (-8 pp)\\
    \textbf{Random} & PSR & 0.90 (-11 pp) & 0.95 (-5 pp) & 0.95 (-5 pp)\\
    \textbf{Random} & GPT & 0.76 (\largest{-24 pp}) & 0.84 (\largest{-16 pp}) & 0.84 (\largest{-16 pp})\\
    \textbf{Random} & GPT+Genre & 0.76 (\largest{-24 pp}) & 0.84 (\largest{-16 pp}) & 0.84 (\largest{-16 pp})\\
    \bottomrule
  \end{tabular}
\end{wraptable}

\paragraph{\textbf{Flip Rates and Overlapping Flips}}
\label{subsec:flip_rates}

To estimate how many AIGTs are relabeled as \emph{human} after rewriting, we track \emph{flip rates} for each combination of explainability method, rewriting strategy, and model. A higher flip rate shows that the corresponding model is more easily deceived by that particular rewriting approach. In these experiments, LIME-based edits coupled with HSR lead to notably high misclassification rates in simpler models such as XGBoost (flipping up to 65\% of the samples), whereas XLM-RoBERTa remains relatively robust at a mere 2.5\% flip rate. GPT-based rewrites result in mid-level flip percentages overall; for example, under LIME+GPT, XGBoost flips 31.67\% of texts while BERT-base and DistilBERT are both at around 9.17\%. Moreover, SHAP+PSR frequently produces lower flip rates, showing that while part-of-speech-constrained edits can deceive less sophisticated classifiers, advanced or ensemble approaches remain relatively unaffected (with an Ensemble flip rate of only 2.5\%). Notably, XLM-RoBERTa’s tendency to overpredict \emph{AI} keeps its flips artificially low. The Ensemble, although not entirely immune, stays in the lower to mid-range across all scenarios, once again showing greater robustness than single-model detectors. Table~\ref{tab:flip_rates_split} provides a detailed breakdown of which sets of models flip each sample in various scenarios.

\begin{table}[ht]
\centering
\scriptsize
\renewcommand{\arraystretch}{1.15}    
\setlength{\tabcolsep}{4pt}          

\caption{Flip rates (\%) from \textbf{AI} to \textbf{human}, grouped by rewriting strategies. 
         Each scenario has 120 AI-labeled items; higher values indicate greater susceptibility to ``flipping.''}
\label{tab:flip_rates_split}

\begin{subtable}[t]{0.48\textwidth}
    \centering
    \caption{HSR \& PSR Strategies}
    \label{tab:flip_hsr_psr}
    \begin{tabular}{
        l
        l
        l
        S[table-format=3.0]
        S[table-format=2.0]
        S[table-format=3.2]
    }
    \toprule
    \textbf{Explain} & \textbf{Strategy} & \textbf{Model} 
      & {\#AI} & {Flips} & {Flip Rate} \\
    \midrule
    \multirow{10}{*}{SHAP}
     & \multirow{5}{*}{HSR}
       & XGBoost      & 120 & 17 & 14.17 \\
     & & BERT         & 120 &  7 &  5.83 \\
     & & DistilBERT   & 120 &  9 &  7.50 \\
     & & XLM-R        & 120 &  1 &  0.83 \\
     & & Ensemble     & 120 &  8 &  6.67 \\
    \cmidrule(lr){2-6}
     & \multirow{5}{*}{PSR}
       & XGBoost      & 120 &  7 &  5.83 \\
     & & BERT         & 120 &  4 &  3.33 \\
     & & DistilBERT   & 120 &  4 &  3.33 \\
     & & XLM-R        & 120 &  0 &  0.00 \\
     & & Ensemble     & 120 &  3 &  2.50 \\
    \midrule
    \multirow{10}{*}{LIME}
     & \multirow{5}{*}{HSR}
       & XGBoost      & 120 & 78 & 65.00 \\
     & & BERT         & 120 & 31 & 25.83 \\
     & & DistilBERT   & 120 & 32 & 26.67 \\
     & & XLM-R        & 120 &  3 &  2.50 \\
     & & Ensemble     & 120 & 33 & 27.50 \\
    \cmidrule(lr){2-6}
     & \multirow{5}{*}{PSR}
       & XGBoost      & 120 & 46 & 38.33 \\
     & & BERT         & 120 & 20 & 16.67 \\
     & & DistilBERT   & 120 & 19 & 15.83 \\
     & & XLM-R        & 120 &  5 &  4.17 \\
     & & Ensemble     & 120 & 19 & 15.83 \\
    \midrule
    \multirow{10}{*}{Random}
     & \multirow{5}{*}{HSR}
       & XGBoost      & 120 & 15 & 12.50 \\
     & & BERT         & 120 & 20 & 16.67 \\
     & & DistilBERT   & 120 & 22 & 18.33 \\
     & & XLM-R        & 120 &  2 &  1.67 \\
     & & Ensemble     & 120 & 20 & 16.67 \\
    \cmidrule(lr){2-6}
     & \multirow{5}{*}{PSR}
       & XGBoost      & 120 & 10 &  8.33 \\
     & & BERT         & 120 & 17 & 14.17 \\
     & & DistilBERT   & 120 & 11 &  9.17 \\
     & & XLM-R        & 120 &  1 &  0.83 \\
     & & Ensemble     & 120 & 14 & 11.67 \\
    \bottomrule
    \end{tabular}
\end{subtable}
\hfill
\begin{subtable}[t]{0.48\textwidth}
    \centering
    \caption{GPT \& GPT+Genre Strategies}
    \label{tab:flip_gpt_gptg}
    \begin{tabular}{
        l
        l
        l
        S[table-format=3.0]
        S[table-format=2.0]
        S[table-format=3.2]
    }
    \toprule
    \textbf{Explain} & \textbf{Strategy} & \textbf{Model} 
      & {\#AI} & {Flips} & {Flip Rate} \\
    \midrule
    \multirow{10}{*}{SHAP}
     & \multirow{5}{*}{GPT}
       & XGBoost      & 120 &  4 &  3.33 \\
     & & BERT         & 120 &  7 &  5.83 \\
     & & DistilBERT   & 120 &  7 &  5.83 \\
     & & XLM-R        & 120 &  0 &  0.00 \\
     & & Ensemble     & 120 & 10 &  8.33 \\
    \cmidrule(lr){2-6}
     & \multirow{5}{*}{GPT+Genre}
       & XGBoost      & 120 &  4 &  3.33 \\
     & & BERT         & 120 &  3 &  2.50 \\
     & & DistilBERT   & 120 &  6 &  5.00 \\
     & & XLM-R        & 120 &  0 &  0.00 \\
     & & Ensemble     & 120 & 11 &  9.17 \\
    \midrule
    \multirow{10}{*}{LIME}
     & \multirow{5}{*}{GPT}
       & XGBoost      & 120 & 38 & 31.67 \\
     & & BERT         & 120 & 11 &  9.17 \\
     & & DistilBERT   & 120 & 11 &  9.17 \\
     & & XLM-R        & 120 &  0 &  0.00 \\
     & & Ensemble     & 120 & 11 &  9.17 \\
    \cmidrule(lr){2-6}
     & \multirow{5}{*}{GPT+Genre}
       & XGBoost      & 120 & 41 & 34.17 \\
     & & BERT         & 120 &  4 &  3.33 \\
     & & DistilBERT   & 120 &  7 &  5.83 \\
     & & XLM-R        & 120 &  0 &  0.00 \\
     & & Ensemble     & 120 &  9 &  7.50 \\
    \midrule
    \multirow{10}{*}{Random}
     & \multirow{5}{*}{GPT}
       & XGBoost      & 120 & 16 & 13.33 \\
     & & BERT         & 120 &  6 &  5.00 \\
     & & DistilBERT   & 120 &  6 &  5.00 \\
     & & XLM-R        & 120 &  0 &  0.00 \\
     & & Ensemble     & 120 &  5 &  4.17 \\
    \cmidrule(lr){2-6}
     & \multirow{5}{*}{GPT+Genre}
       & XGBoost      & 120 & 13 & 10.83 \\
     & & BERT         & 120 &  6 &  5.00 \\
     & & DistilBERT   & 120 &  9 &  7.50 \\
     & & XLM-R        & 120 &  0 &  0.00 \\
     & & Ensemble     & 120 & 11 &  9.17 \\
    \bottomrule
    \end{tabular}
\end{subtable}

\end{table}

As shown in Table~\ref{tab:flip_rates_split}, many rewriting strategies cause a lot of flips for some detectors, especially the simpler ones. However, the Ensemble model usually has a lower flip rate. This shows that using several models together makes detection more reliable.

\begin{figure*}[ht]
\centering

\includegraphics[width=0.24\textwidth]{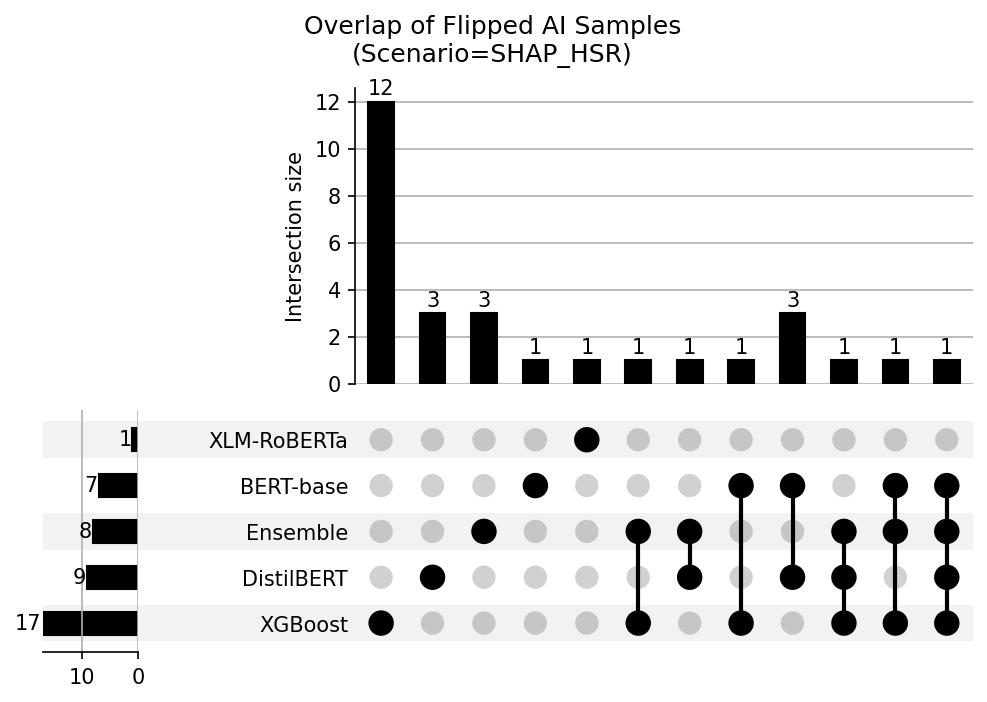}
\includegraphics[width=0.24\textwidth]{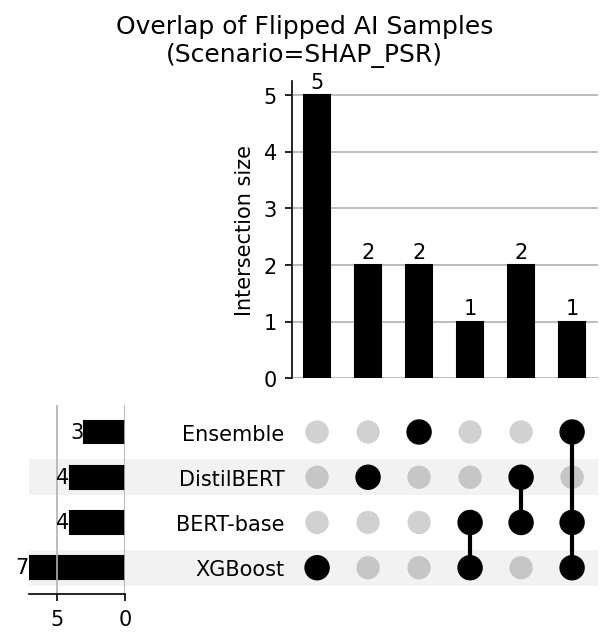}
\includegraphics[width=0.24\textwidth]{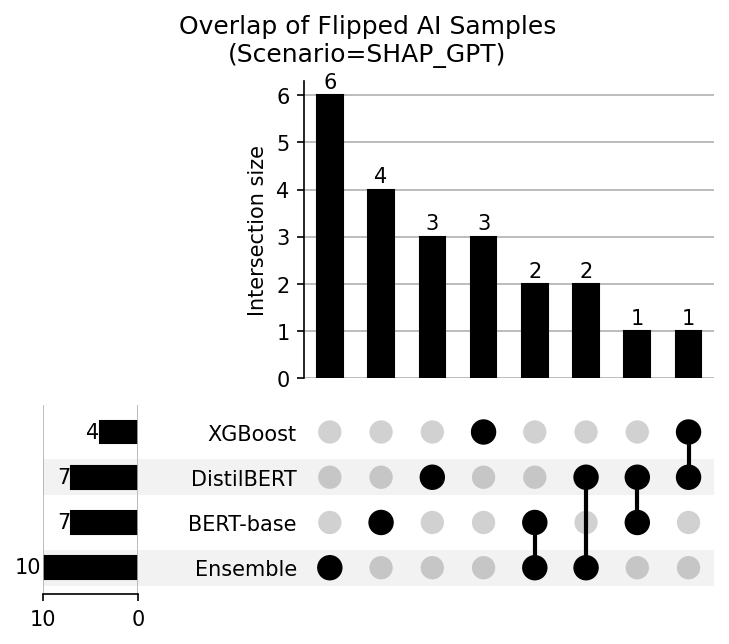}
\includegraphics[width=0.24\textwidth]{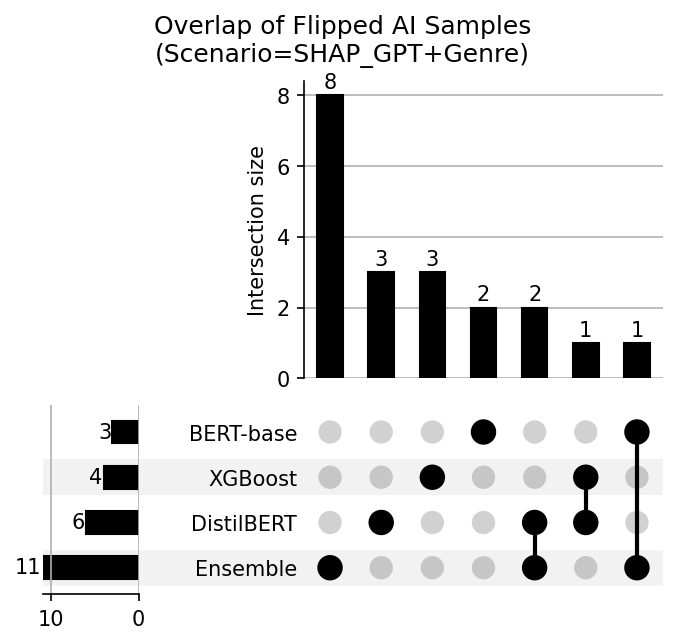}

\includegraphics[width=0.24\textwidth]{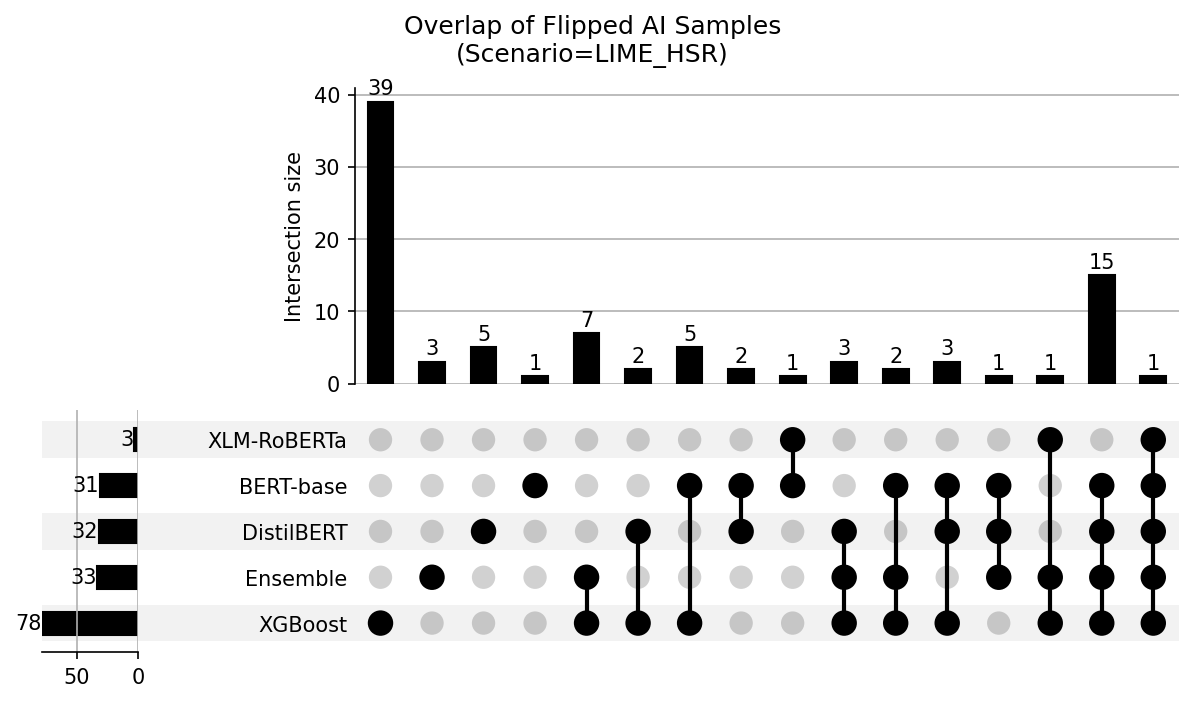}
\includegraphics[width=0.24\textwidth]{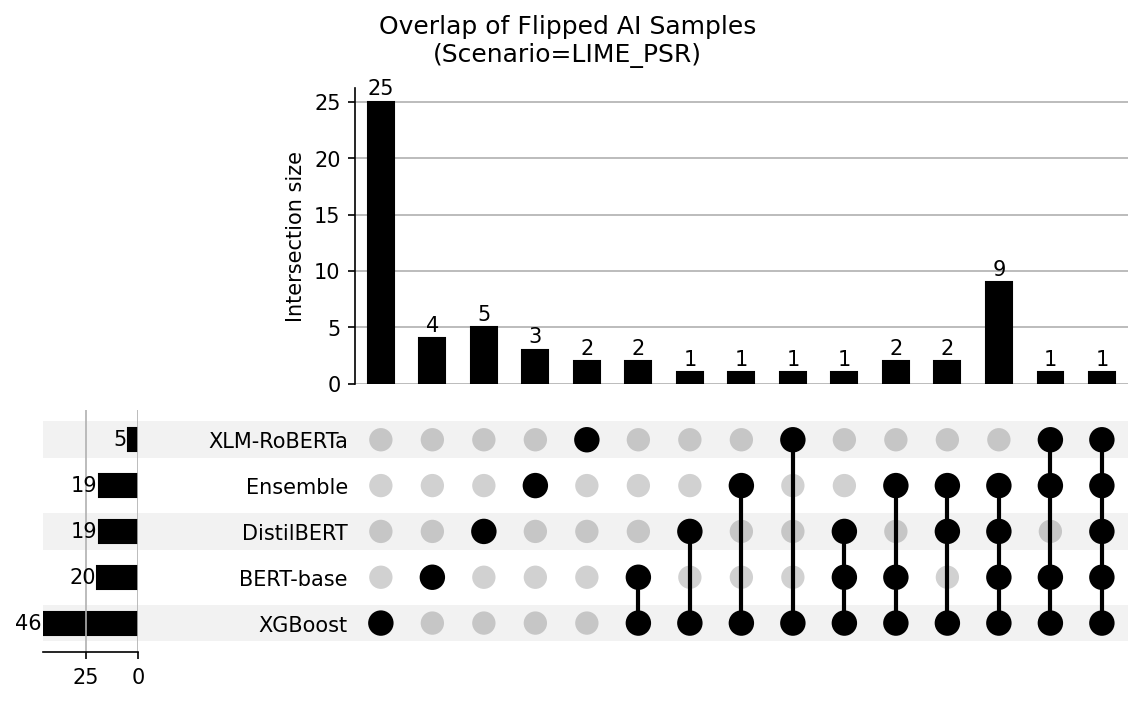}
\includegraphics[width=0.24\textwidth]{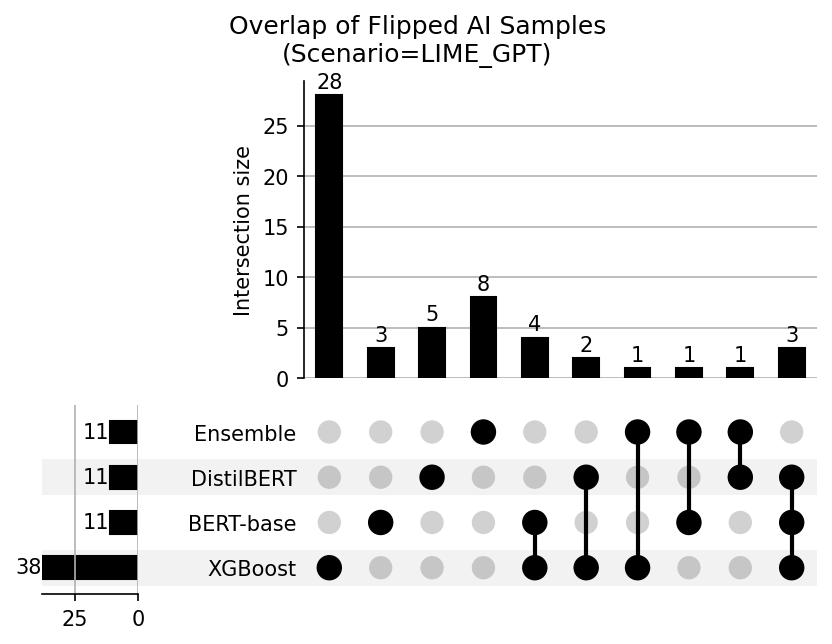}
\includegraphics[width=0.24\textwidth]{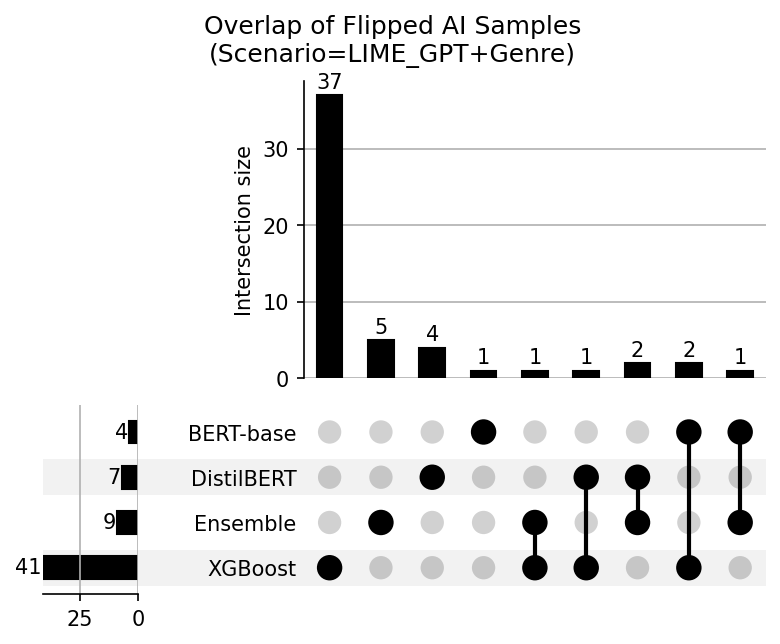}

\includegraphics[width=0.24\textwidth]{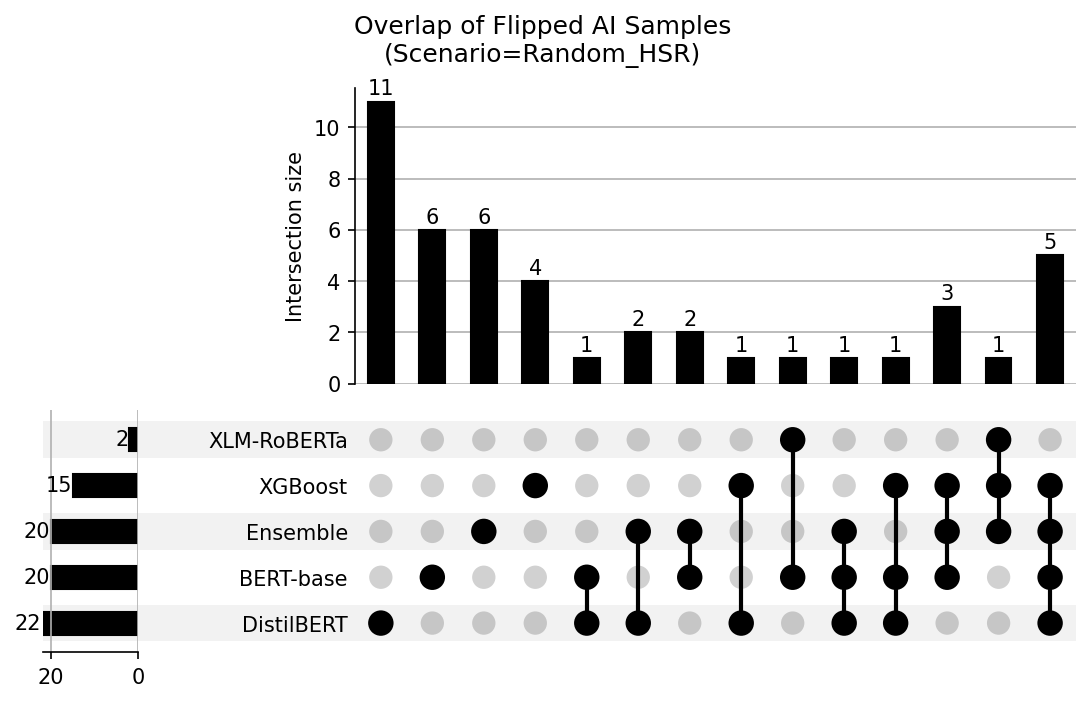}
\includegraphics[width=0.24\textwidth]{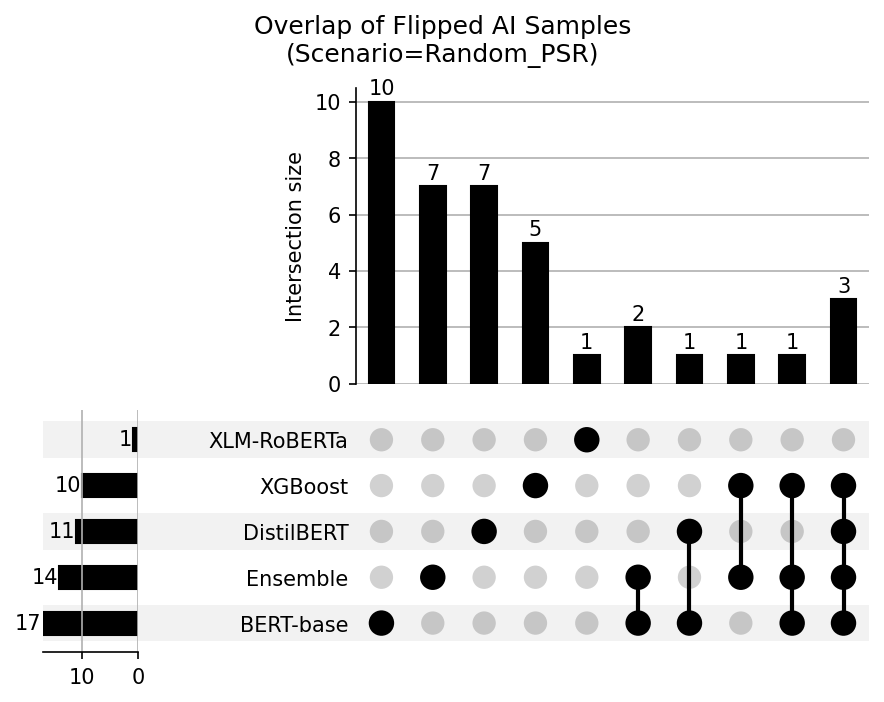}
\includegraphics[width=0.24\textwidth]{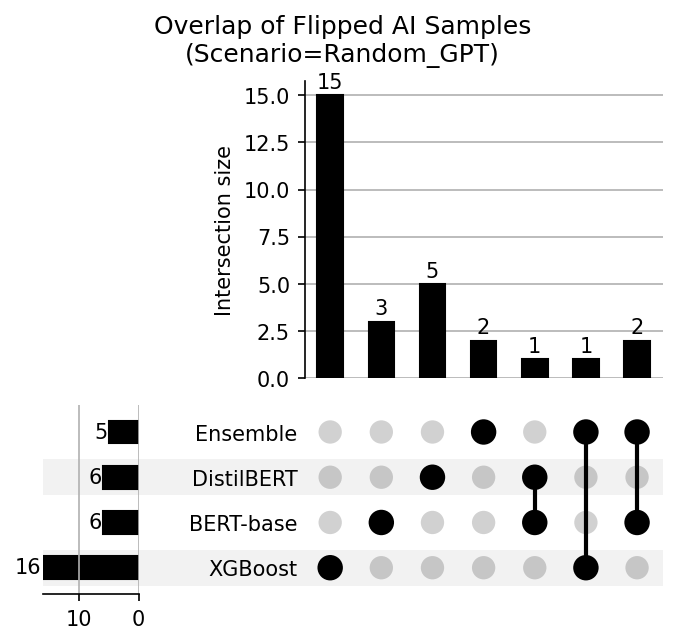}
\includegraphics[width=0.24\textwidth]{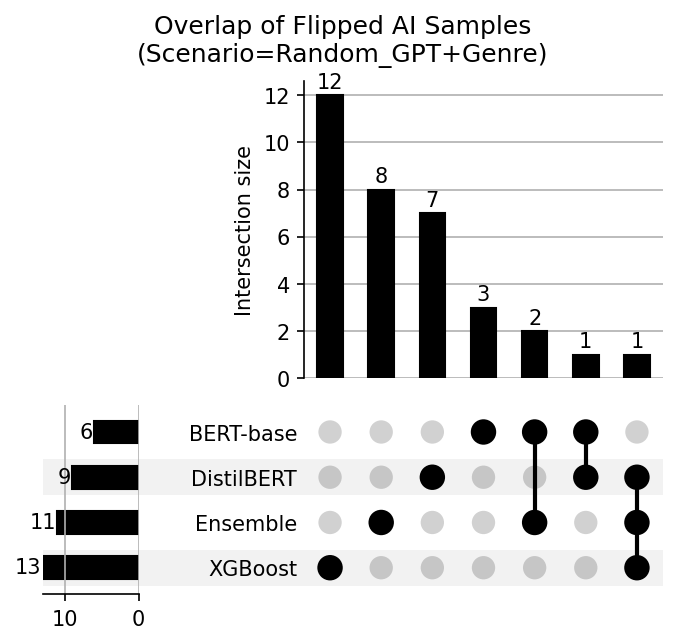}

\caption{\small Overlaps among flipped AI samples under different explainability methods and rewriting strategies. Each bar corresponds to the number of samples simultaneously flipped by one or more models.}
\label{fig:all_upset_plots}
\end{figure*}

\paragraph{\textbf{Interpreting Overlap Patterns.}}
Figure~\ref{fig:all_upset_plots} visualizes the overlap patterns among flipped samples across all models using UpSet plots. Each bar represents the number of AI-generated samples that were simultaneously flipped (misclassified as human) by a specific combination of models, with the connected dots below indicating which models are involved in that intersection.

Several patterns emerge from this analysis. First, XGBoost-only flips dominate many scenarios, particularly under LIME+HSR where XGBoost uniquely flips 40+ samples that other models correctly classify. This indicates that XGBoost's feature-based approach is most susceptible to targeted token replacements. Second, the Ensemble exhibits notably smaller exclusive flip counts compared to other models---typically 3--11 samples depending on the scenario---demonstrating that samples fooling the Ensemble are usually also fooling other models, suggesting these are inherently difficult cases rather than ensemble-specific weaknesses.

Third, the overlap between XGBoost and transformer models is relatively limited, confirming that their failure modes differ substantially. This complementarity is precisely why the ensemble architecture succeeds: when one detection pathway fails, others often succeed. For instance, under SHAP+GPT, fewer than 5 samples fool all five models simultaneously, meaning the vast majority of evasion attempts fail against at least one component.

Fourth, XLM-RoBERTa shows consistently minimal participation in flip intersections due to its tendency to overpredict AI, which paradoxically makes it resistant to evasion attacks while reducing precision. The Ensemble balances this trade-off by combining XLM-RoBERTa's robustness with other models' higher precision.

These overlap patterns quantify the ensemble's robustness advantage: even when 27.5\% of samples flip under LIME+HSR for the Ensemble, the shared samples with other models suggest these represent the hardest adversarial cases where extensive token modifications fundamentally alter the text's detectable characteristics

In general, GPT-based replacements make text more natural or domain-specific, which greatly impacts models that rely on simple token patterns for detection. HSR and PSR (especially with SHAP/LIME) can also be effective while keeping a higher BLEU score, meaning fewer changes. Since PSR aligns part-of-speech tags, the text remains fluent but loses patterns AI might use, leading to performance drops in classifiers like XLM-RoBERTa and XGBoost. In contrast, our Ensemble model is more robust, as its combined transformer knowledge helps it detect AI content even after targeted token edits. Overall, our results show that the Ensemble model is the most accurate and resilient, effectively handling English and Dutch texts across different genres. While rewriting reduces detection rates, it does not fully bypass the Ensemble model.

\subsection{Analysis of Ensemble Components}
\label{subsec:ablation}

To understand the contribution of individual components to our ensemble's performance, we analyze the results from Table~\ref{tab:lang_domain_updated}, which shows each single model's performance alongside the Ensemble. This analysis serves as a proxy for a leave one component out analysis, revealing how much each component contributes to the combined system.

\paragraph{Ensemble vs. Single Models.}
Across all language-domain combinations, the Ensemble consistently outperforms or matches the best single model. In English, the Ensemble achieves an average F1 of 0.86 compared to 0.81 for BERT-base, 0.84 for DistilBERT, and 0.68 for XLM-RoBERTa. In Dutch, the performance gap is even more pronounced: the Ensemble averages 0.78 F1 while BERT-base achieves only 0.60, DistilBERT 0.65, and XLM-RoBERTa 0.63. This demonstrates that the ensemble architecture provides substantial gains, particularly for challenging cross-lingual scenarios.

\paragraph{Complementary Strengths.}
Our ensemble benefits from the complementary capabilities of its constituent models. XLM-RoBERTa provides strong multilingual representations, particularly valuable for Dutch text. BERT-base excels in English news and reviews where domain-specific patterns are important. DistilBERT offers computational efficiency while maintaining competitive performance. The frozen branch (pre-trained on AuTexTification) captures general AI-detection patterns, while the fresh branch (fine-tuned on augmented CLIN33) adapts to domain-specific characteristics. This dual-branch architecture enables the ensemble to leverage both broad knowledge transfer and targeted domain adaptation.

\paragraph{Robustness Under Adversarial Conditions.}
Table~\ref{tab:all_models_rewriting_f1acc} reveals that the Ensemble exhibits smaller performance drops under token replacement attacks compared to single models. While XGBoost shows drops of up to 56 percentage points (from 0.81 to 0.25 F1) under LIME+HSR, the Ensemble's maximum drop is 19 percentage points (from 0.83 to 0.64), demonstrating enhanced robustness. This resilience stems from the ensemble's redundant detection pathways---when one component is fooled by specific token manipulations, others may still correctly identify the AI-generated nature of the text.

\paragraph{Limitations.}
A complete leave one component out analysis would require retraining the ensemble with each component removed, which was beyond the scope of our current experimental setup. Future work could systematically evaluate the marginal contribution of each model by training multiple ensemble variants. Nevertheless, the single-model results provide strong evidence that the ensemble architecture meaningfully combines component strengths rather than merely averaging their predictions.

\subsection{Sensitivity Analysis}
\label{subsec:sensitivity}

Our token replacement framework raises questions about the relationship between the number of tokens modified and the resulting detection evasion. While a systematic parameter sweep across different token counts was not conducted in this study, we can draw insights from the observed patterns.

\paragraph{Fidelity-Evasion Trade-off.}
The similarity metrics in Table~\ref{tab:rewriting_similarity} reveal a clear trade-off between textual fidelity and evasion effectiveness. PSR-based methods achieve the highest BLEU scores (0.97 for SHAP+PSR), indicating minimal text modification, while producing moderate flip rates. In contrast, GPT-based methods show lower BLEU scores (0.76--0.86 depending on explainability method), reflecting more substantial rewrites, but do not necessarily achieve higher flip rates against the Ensemble. This suggests diminishing returns: beyond a certain modification threshold, additional changes may not improve evasion and could potentially introduce new AI-detectable patterns.

\paragraph{Influence of Token Selection Strategy.}
Comparing SHAP, LIME, and Random token selection reveals that targeting influential tokens (as identified by XAI methods) produces larger detection drops with fewer modifications. Under the same replacement strategy, LIME-selected tokens generally cause greater performance degradation than random selection, confirming that explainability-guided attacks are more efficient than random perturbations. The flip rates in Table~\ref{tab:flip_rates_split} show that LIME+HSR achieves higher evasion against XGBoost with similar textual changes compared to Random+HSR.

\paragraph{Token Count Considerations.}
Our experiments replaced the top-K most influential tokens as identified by SHAP or LIME, where K was determined by the importance threshold rather than a fixed count. Based on prior work~\citep{zhou2024humanizing}, we expect that most detection signal concentrates in the top 5--15\% of tokens. Future work could systematically vary K to characterize the sensitivity curve and identify optimal modification levels that balance evasion success against textual degradation.

\subsection{Human Evaluation}
\label{subsec:human_eval}

To complement automated metrics, we conducted a small-scale human evaluation to assess the perceptual quality of rewritten texts and human ability to detect AI-generated content.

\paragraph{Setup.} We selected 100 text samples stratified by rewriting strategy and model flip patterns: samples from each major rewriting condition (HSR, PSR, GPT, GPT+Genre), plus original AI texts and human-written controls from the same domains. The distribution included 58 samples that successfully evaded detection (``flipped'': 18 HSR, 14 PSR, 14 GPT, 12 GPT+Genre), 12 that remained detected despite rewriting, 15 unmodified AI-generated texts, and 15 human-written texts. Samples were selected to represent diverse scenarios from our overlap analysis, including cases where multiple models were fooled and cases where only specific models were affected.

Two evaluators, blind to the experimental condition, rated each text on three dimensions: (1) \textit{Fluency} (1--5 scale: how natural does the text read?), (2) \textit{Coherence} (1--5 scale: is the text logically consistent?), and (3) \textit{AI Detection} (5-point scale from ``Definitely Human'' to ``Definitely AI''). We report inter-rater reliability using Cohen's $\kappa$.

\paragraph{Results.} Table~\ref{tab:human_eval} summarizes the human evaluation findings.

\begin{table}[ht]
\centering
\caption{Human Evaluation Results (N=100 samples, 2 evaluators). Fluency and Coherence are rated on 1--5 scales. Detection shows the percentage of correct AI/human classifications by human evaluators. Cohen's $\kappa$ for inter-rater reliability: Fluency=0.12, Coherence=0.14, Detection=0.08.}
\label{tab:human_eval}
\begin{tabular}{lcccc}
\toprule
\textbf{Condition} & \textbf{N} & \textbf{Fluency} & \textbf{Coherence} & \textbf{Detection} \\
 & & (1-5) & (1-5) & (\% Correct) \\
\midrule
Original AI & 15 & 4.1 ($\pm$0.8) & 4.2 ($\pm$0.6) & 56.7\% \\
Human Control & 15 & 4.4 ($\pm$0.6) & 4.6 ($\pm$0.5) & 63.3\% \\
Rewritten (Flipped) & 58 & 3.9 ($\pm$0.8) & 4.0 ($\pm$0.9) & 37.9\% \\
Rewritten (Detected) & 12 & 3.8 ($\pm$0.6) & 3.6 ($\pm$0.7) & 58.3\% \\
\midrule
\textit{Overall} & 100 & 4.0 & 4.1 & 47.0\% \\
\bottomrule
\end{tabular}
\end{table}

\paragraph{Observations.} The human evaluation reveals several key findings. First, human evaluators achieved only 47.0\% overall detection accuracy, substantially lower than the Ensemble model's 83\% accuracy on the full held-out test set, suggesting that machine detectors currently outperform average human readers at identifying AI-generated text. Second, texts that successfully evaded machine detection (``Flipped'' condition) also fooled humans at high rates, with only 37.9\% correctly identified as AI-generated---compared to 56.7\% for original AI texts. This correlation between machine and human vulnerability suggests that our token replacement strategies effectively humanize AI text along dimensions that both humans and machines rely on for detection. Third, all conditions maintained high fluency (3.8--4.4) and coherence (3.6--4.6) ratings, indicating that rewriting strategies preserve text quality while reducing detectability. The modest inter-rater reliability ($\kappa \approx 0.11$) reflects the inherent difficulty of AI detection for non-expert readers, consistent with findings from prior human evaluation studies in AI-generated text detection~\citep{clark2021all}.

\section{Discussion and Conclusion}
\label{sec:discussion_conclusion}
Our findings show that explanation techniques such as SHAP and LIME can guide systematic token replacements that make AIGT less likely to be classified as \emph{AI-generated}. When we identify and edit tokens most indicative of artificial content, simpler models like XGBoost can be misled at high rates, especially under LIME-based rewrites combined with synonym substitutions. Yet even more advanced models, such as BERT-base or DistilBERT, still show vulnerability in certain scenarios, particularly when GPT-generated replacements transform the text more extensively. These methods, however, do not completely deceive our Ensemble model, which remains comparatively robust across languages, genres, and rewriting methods.

\paragraph{On the Significance of Performance Differences.} While the absolute performance differences between the Ensemble and individual models may appear modest in isolation, several factors underscore their practical importance. First, these differences are consistent across diverse experimental settings (Section~\ref{subsec:setup}), with the Ensemble reliably outperforming baselines across languages, domains, and adversarial strategies. Second, when deployed at scale---processing thousands or millions of texts daily in real-world content moderation scenarios---even small percentage improvements translate to substantial reductions in false positives and negatives. Third, and perhaps most importantly, the Ensemble's primary advantage lies not in raw accuracy but in its \textit{consistency}: it exhibits lower variance across languages, domains, and adversarial strategies than any single model. This robustness property is crucial for practical deployment, where unpredictable failure modes pose greater risks than marginally lower average performance.

\paragraph{Contributions and Novelty.} This work makes several distinct contributions to the growing field of AI-generated text detection. To our knowledge, this is the first systematic study to leverage explainability methods for both adversarial evasion \textit{and} robust detection within a unified framework. While prior work has explored either XAI-guided attacks~\cite{david2025authormist} or detection systems in isolation, our dual perspective provides unique insights into the arms race between detection and evasion. Additionally, our frozen-plus-fresh ensemble architecture represents a novel design that balances the stability of pre-trained representations with domain-specific adaptability. Our cross-lingual (English and Dutch), multi-domain (news, reviews, social media) evaluation further distinguishes this work from studies limited to English-only or single-domain settings. Finally, the detailed analysis of token replacement strategies---from conservative synonym substitution to aggressive LLM-based rewriting---offers actionable guidance for both detection system designers and policymakers concerned with AI content transparency.

From the perspective of text similarity, we see that methods like HSR and PSR typically achieve higher BLEU and ROUGE scores, indicating minimal alterations to the original text. This can preserve meaning and style, but targeted token swaps can still hide enough AI signals to disrupt certain detectors. In contrast, GPT-based rewrites, especially GPT+Genre, often provide more comprehensive revisions, making the text harder to classify but also changing more from the source. Although such extensive rewriting can produce higher flip rates, our Ensemble classifier still maintains better overall performance than single-model systems.

In practice, these results underscore both the potential and the limitations of adversarial rewriting. Minor edits guided by SHAP or LIME can undermine certain classifiers. Future work might extend to more sophisticated paraphrasing of entire sentences or incorporate style transfer models that better replicate a specific human author. At the same time, it is crucial to consider watermarking and policy frameworks that ensure transparency when AIGT is involved. Our experiments show that even when AI content is skillfully masked, detection remains possible with multi-architecture approaches.

\paragraph{Limitations and Future Directions.} Several limitations of this study merit acknowledgment. First, the CLIN33 dataset, while carefully curated, is relatively modest in size; larger-scale evaluations would strengthen confidence in our findings. Second, the AI-generated content in our datasets was produced by GPT-4 and Vicuña-13B for CLIN33, and GPT-3 and BLOOM-series models for the AuTexTification subset, which represent an earlier generation of LLMs. The detection landscape is evolving rapidly, and content from newer models such as Claude, Llama 3, and their variants may exhibit different patterns that could affect both detection accuracy and adversarial vulnerability. Third, our evaluation covers two languages (English and Dutch) and three domains; generalization to other languages, particularly those with different morphological structures, and to specialized domains such as scientific writing or legal text remains to be demonstrated. Fourth, while we provide computational cost estimates, we did not conduct a systematic study of the efficiency-accuracy trade-off under deployment constraints. Future work should address these gaps by: (1) evaluating on larger, more diverse datasets including content from the latest LLMs; (2) exploring additional languages and domains; (3) investigating combinations of XAI-guided rewriting with other adversarial techniques; (4) developing adaptive detection systems that can co-evolve with increasingly sophisticated evasion methods; and (5) extending human evaluation with larger, more diverse annotator pools including domain experts and professional content moderators, which would provide deeper insights into the aspects that distinguish AI-generated from human-written text.

In conclusion, our study shows how explanation methods can inform token replacements to disguise AIGT, yet also reveals that robust ensemble classifiers are harder to deceive. Balancing improved detection methods with the need for responsible use of writing aids will remain an important challenge. We hope these findings contribute to a deeper understanding of both the potential vulnerabilities in existing AI detection models and the methods that can keep them resilient against adversarial transformations.

\begin{acks}
We thank the CLIN33 shared task organizers for making the dataset available, and we appreciate the helpful feedback from colleagues who guided our research directions. We also thank SURF for providing the computational facilities. We are grateful to Tina Shahedi for her contribution to the human evaluation study. 

\end{acks}

\bibliographystyle{ACM-Reference-Format}
\bibliography{01-sample-base}

 \appendix

\clearpage
\section*{Reproducibility Checklist for JAIR}

\noindent\textbf{All articles:}
\begin{enumerate}
    \item \textbf{All claims investigated in this work are clearly stated.} \\
    \textit{Answer:} Yes, we clearly state our main claims: we want to see how effective rewriting with SHAP/LIME is, and how an ensemble model performs.
    
    \item \textbf{Clear explanations are given how the work reported substantiates the claims.} \\
    \textit{Answer:} Yes, we explain how each experiment’s results support our claims.
    
    \item \textbf{Limitations or technical assumptions are stated clearly and explicitly.} \\
    \textit{Answer:} Some. We focus on detection and rewriting only, not on other possible aspects.
    
    \item \textbf{Conceptual outlines and/or pseudo-code descriptions of the AI methods introduced in this work are provided, and important implementation details are discussed.} \\
    \textit{Answer:} Yes, our method flow is clearly described, especially for the ensemble approach in the algorithm~\ref{fig:ms-structure}.
    
    \item \textbf{Motivation is provided for all design choices, including algorithms, implementation choices, parameters, data sets, and experimental protocols beyond metrics.} \\
    \textit{Answer:} Yes, we explain why we use these models, datasets, and rewriting strategies.
\end{enumerate}

\subsection*{Articles containing theoretical contributions}
\noindent\textit{Does this paper make theoretical contributions?} \\
\textit{Answer:} No. (We do not do formal theoretical proofs.)

\subsection*{Articles reporting on computational experiments}
\noindent\textit{Does this paper include computational experiments?} \\
\textit{Answer:} Yes.
\begin{enumerate}
    \item \textbf{All source code required for experiments is included or will be made publicly available upon publication, following best practices.} \\
    \textit{Answer:} We plan to share the main code base (e.g., on GitHub).
    
    \item \textbf{The source code comes with a license that allows free usage for reproducibility purposes.} \\
    \textit{Answer:} Yes, we use a permissive license.
    
    \item \textbf{The source code comes with a license that allows free usage for research purposes in general.} \\
    \textit{Answer:} Yes.
    
    \item \textbf{Raw, unaggregated data from all experiments is included or will be made publicly available.} \\
    \textit{Answer:} We will share our final splits, but some augmented data might be partial.
    
    \item \textbf{The unaggregated data comes with a license that allows free usage for reproducibility purposes.} \\
    \textit{Answer:} Yes, as far as allowed by the original dataset licenses.
    
    \item \textbf{The unaggregated data comes with a license that allows free usage for research purposes in general.} \\
    \textit{Answer:} Yes, within the limits of the original dataset licenses.
    
    \item \textbf{If an algorithm depends on randomness, the method used for generating random numbers and setting seeds is described sufficiently.} \\
    \textit{Answer:} We set random seeds but do not deeply document random generation details.
    
    \item \textbf{The execution environment for experiments (hardware/software) is described.} \\
    \textit{Answer:} Yes, we mention key libraries (e.g., Hugging Face Transformers, PyTorch).
    
    \item \textbf{The evaluation metrics used are clearly explained and motivated.} \\
    \textit{Answer:} Yes, we use F1, accuracy, BLEU, and ROUGE and explain why.
    
    \item \textbf{The number of algorithm runs used to compute each result is reported.} \\
    \textit{Answer:} Each experiment is run with 4 random seeds; we report the mean (standard deviation not reported in tables).
    
    \item \textbf{Reported results have not been “cherry-picked.”} \\
    \textit{Answer:} Yes, we show both successes and failures.
    
    \item \textbf{Analysis of results goes beyond single-dimensional summaries of performance.} \\
    \textit{Answer:} We show F1, accuracy, and text-similarity. (We do not always show standard deviation or confidence intervals.)
    
    \item \textbf{All (hyper-) parameter settings for the algorithms/methods used are reported, along with the rationale or method for determining them.} \\
    \textit{Answer:} Yes, we provide learning rates, batch sizes, etc.
    
    \item \textbf{The number and range of (hyper-) parameter settings explored prior to final experiments have been indicated.} \\
    \textit{Answer:} We mention random search, though not each detail.
    
    \item \textbf{Appropriately chosen statistical hypothesis tests are used to establish significance in the presence of noise.} \\
    \textit{Answer:} We do not do formal significance testing.
\end{enumerate}

\subsection*{Articles using data sets}
\noindent\textit{Does this work rely on one or more data sets?} \\
\textit{Answer:} Yes.

\begin{enumerate}
    \item \textbf{All newly introduced data sets are included in an online appendix or will be made publicly available with a suitable license.} \\
    \textit{Answer:} We mostly use public sets (CLIN33, AuTexTification) and will release our augmented version.
    
    \item \textbf{The newly introduced data set comes with a license that allows free usage for reproducibility purposes.} \\
    \textit{Answer:} Yes, for the augmented subset.
    
    \item \textbf{The newly introduced data set comes with a license that allows free usage for research purposes in general.} \\
    \textit{Answer:} Yes.
    
    \item \textbf{All data sets drawn from the literature or other public sources are accompanied by appropriate citations.} \\
    \textit{Answer:} Yes, we cite the original authors.
    
    \item \textbf{All data sets drawn from the existing literature are publicly available.} \\
    \textit{Answer:} Yes, with possible constraints.
    
    \item \textbf{All new or non-public data sets are described in detail, including relevant statistics, collection, and annotation processes.} \\
    \textit{Answer:} Yes, for our augmented data.
    
    \item \textbf{All methods used for preprocessing, augmenting, batching, or splitting data sets are described in detail.} \\
    \textit{Answer:} Yes, we detail how we clean and split.
\end{enumerate}

\subsection*{Explanations on any of the answers above (optional):}
We use mostly public data and provide augmented sets where possible. Some data licensing is inherited from the original corpora. Where possible, we provide reproducible code, scripts, and subsets to ensure others can replicate our results.

\end{document}